\documentclass{article}
\usepackage{ijcai26}

\usepackage{times}
\usepackage{soul}
\usepackage{url}
\usepackage[hidelinks]{hyperref}
\usepackage[utf8]{inputenc}
\usepackage[small]{caption}
\usepackage{graphicx}
\usepackage{amsmath}
\usepackage{amsthm}
\usepackage{booktabs}
\usepackage{algorithm}
\usepackage{algorithmic}
\usepackage[switch]{lineno}
\usepackage{caption}
\usepackage{subcaption}

\usepackage{xcolor}
\usepackage{pifont}

\usepackage[round,authoryear]{natbib}
\usepackage{amsmath,amssymb,amsfonts}

\title{Beyond Accuracy: Evaluating Efficiency, Robustness and Explainability in Deep Learning for Malaria Diagnosis}

\author{
Olivier Kanamugire$^1$
\and
Kerol Djoumessi$^{2}$\and
\affiliations
$^1$African Institute for Mathematical Sciences, Kigali, Rwanda\\
$^2$Hertie Institute for AI in Brain Health, University of T\unexpanded{\"u}bingen, Germany \\
\emails
olivier.kanamugire@aims.ac.rw,
kerol.djoumessi-donteu@uni-tuebingen.de
}

\begin{document}

\maketitle

\begin{abstract}
    Malaria remains a leading cause of mortality in sub-Saharan Africa, where scarce diagnostic infrastructure makes timely, accurate diagnosis particularly challenging. While deep learning offers a compelling path toward automated malaria screening, clinical adoption is hindered by computational cost and opacity in decision-making. This work benchmarks four deep learning models spanning a wide range of designed design architectures and model capacities on the NLM-Malaria dataset, jointly evaluating predictive performance, robustness, and post-hoc explainability. We find that lightweight, efficient-by-design models match their heavier counterparts in predictive performance, and the Friedman test confirms no statistically significant performance differences. 
    CAM-based XAI methods consistently localize diagnostically relevant regions, while fine-grained attribution methods produce less targeted explanations, particularly with heavier backbones. Robustness evaluation under three types of image corruption further reveals that model confidence degrades faster than accuracy, providing a practical signal for human review. However, no XAI method is robust to corruption, with explanation reliability degrading at noise levels plausible in clinical practice, even when predictions remain accurate. 
    These findings support the deployment of lightweight architectures for malaria diagnosis in resource-constrained settings, while highlighting the vulnerability of post-hoc explanations as an important consideration for responsible clinical deployment.
\end{abstract}


\section{Introduction}
    Malaria remains a serious global health challenge, with an estimated 282 million cases and 610 000 deaths in 2024---roughly 9 million more cases than the previous year~\cite{who2025world}. Total malaria deaths increased by 5.5\% between 2015 and 2024, with over a third of that rise occurring in the single year from 2023 to 2024 \cite{who2025world}.
    The burden is geographically concentrated: the WHO African Region accounts for an estimated 94\% of global cases and 95\% of deaths, with 75\% of fatalities in that region occurring in children under five~\cite{who2025world}. Although global efforts have prevented an estimated 2.3 billion cases and 14 million deaths since 2000, the recent rise in mortality highlights the ongoing need for early and accurate diagnosis, particularly in sub-Saharan African countries.

    The disease is caused by parasites of the genus \textit{Plasmodium}, transmitted to humans through the bites of infected female \textit{Anopheles} mosquitoes~\cite{coatney1971primate, Marcus2009Malaria}. Once in the bloodstream, the parasite multiplies in the liver before invading and destroying red blood cells, producing the disease's characteristic symptoms. Of the more than 200 known \textit{Plasmodium} species, five are recognized to infect humans~\cite{coatney1971primate, Marcus2009Malaria}. \textit{P. falciparum} is the most lethal and widespread, responsible for the majority of malaria-related fatalities. \textit{P. vivax} is globally distributed but predominates in temperate and some tropical regions. \textit{P. ovale} is comparatively rare, with its highest prevalence in equatorial Africa, while \textit{P. malariae} is a cosmopolitan species thriving in hot climates~\cite{coatney1971primate}. The more recently identified \textit{P. knowlesi}, endemic to Southeast Asia, is potentially fatal but responds well to early treatment~\cite{jeremiah2014challenges}.
    The diagnostic gold standard for malaria remains light microscopy of Giemsa-stained blood smears, through which trained microscopists identify parasitized erythrocytes based on morphological changes in cell size, shape, and pigmentation~\cite{fitri2022malaria}. While accurate in well-resourced settings, this approach is labor-intensive and prone to human error. In malaria-endemic regions---particularly across sub-Saharan Africa---the scarcity of trained professionals and diagnostic infrastructure creates a compelling case for automated, scalable detection systems~\cite{wongsrichanalai2007review, maturana2022advances}.

    Deep learning, and convolutional neural networks (CNNs) in particular, have demonstrated strong performance on malaria detection from blood smear imaging, often matching or exceeding expert-level sensitivity and specificity \cite{maturana2022advances, fitri2022malaria}. However, clinical adoption has been constrained by a fundamental limitation: opacity in decision-making \cite{lipton2018mythos}. In high-stakes diagnostic settings, clinicians and regulatory bodies require not only accurate predictions, but also mechanistic justification---an understanding of which image features drove a given classification \cite{samek2017explainable}. The development of interpretable deep learning systems for malaria diagnosis therefore represents both a technical and a translational priority.

    This study investigates four deep learning architectures spanning distinct design and model capacities: ResNet-18 \cite{he2016deep}, EfficientNet-B0 \cite{tan2019efficientnet}, MobileNet-v3-Large \cite{howard2017mobilenets}, and Vision Transformer (ViT-B/16) \cite{dosovitskiy2020image}. ResNet-18 introduces skip connections that enable the network to preserve and propagate information across layers, whereas EfficientNet and MobileNet emphasize computational efficiency through different mechanisms: the former via compound scaling of depth, width, and resolution, and the latter through depthwise separable convolutions that reduce computational complexity. In contrast, ViT adopts a fundamentally different paradigm by replacing convolutions with patch-based self-attention, representing image regions as contextually related token embeddings. 
    Beyond predictive performance, the study further analyzes the post-hoc explanations generated by each model. The primary objective is not solely to determine which architecture achieves the highest performance, but to examine how predictive accuracy, model capacity, and explainability interact, thereby characterizing the trade-off between accuracy, efficiency, and interpretability that is critical for deploying machine learning systems in clinical settings.

\section{Related  Work}
    \paragraph{Transfer learning.} Transfer learning addresses the challenge of limited labeled training data by fine-tuning models pre-trained on large datasets to downstream target tasks, thereby leveraging previously learned feature representations to improve performance and computational efficiency \cite{pan2009survey}. In the context of malaria diagnosis, strong empirical results have been reported. For instance,\cite{ramos2024transfer} evaluated six pre-trained architectures for \textit{P. vivax} detection, with DenseNet201 achieving a mean AUC of $99.41\%$, while \cite{mujahid2024efficient} reported $97.6\%$ classification accuracy using a fine-tuned EfficientNet-B2, outperforming several CNN baselines including VGG-16 and ResNet variants. Nevertheless, transfer learning is not universally optimal;\cite{sabha2024scratch} showed that its effectiveness depends on factors such as dataset size, domain similarity, and task formulation, underscoring the importance of informed model selection over the common adoption of pre-trained architectures.
  
    \paragraph{Explainable AI (XAI).} Interpretability has become increasingly important as machine learning models are deployed in high-stakes domains like clinical diagnosis. Although models may achieve strong benchmark performance, they can still rely on spurious correlations or fail in clinically critical cases---limitations that conventional evaluation metrics alone may not capture \cite{samek2017explainable}. In medical contexts, the lack of reliable explanation can have significant consequences, as incorrect predictions may directly affect patient outcomes. This highlights a fundamental trade-off between model complexity and interpretability: deep neural networks can capture highly subtle patterns through millions of parameters, yet their decision-making processes often remain difficult to interpret \cite{lipton2018mythos}.
    To address this challenge, post-hoc explainability methods such as Grad-CAM \cite{selvaraju2017grad} and SHAP \cite{lundberg2017unified} have been applied to malaria classification tasks. For example,\cite{parveen2025trustworthy} reported that Grad-CAM explanations aligned closely with model confidence, revealing that misclassifications frequently originated from staining artifacts or ambiguous cellular morphology. Similarly, \cite{ahamed2025improving} used Grad-CAM and SHAP to analyze performance differences across CNN architectures, emphasizing the role of interpretability in building clinically reliable systems. However, each explanation method presents important limitations: Grad-CAM typically produces coarse localization maps, and SHAP may exhibit instability under perturbations or often incur substantial computational cost.

    \paragraph{Model efficiency.} Deploying deep learning models in clinical settings presents practical constraints, particularly in low-resource settings where inference may need to operate on resource-limited hardware such as mobile devices or edge-based diagnostic systems with restricted memory, computation, and power budgets \cite{nguyen2026efficient}. Several model compression strategies have been proposed to address these limitations \cite{nguyen2026efficient}, including pruning, which removes low-contribution weights or filters; quantization, which reduces numerical precision of model parameters; and knowledge distillation, which transfers knowledge from a large teacher model to a compact student model. 
    An alternative to post-hoc compression is efficiency by design, where architectural choices inherently reduce computational cost through built-in inductive biases \cite{howard2017mobilenets, tan2019efficientnet}. In this study, this latter approach is considered by focusing on MobileNetV3 and EfficientNet-B0, both designed for lightweight inference while maintaining competitive diagnostic performance.

\section{Methods}
    \subsection{Model architecture}
        \begin{figure}[H]
            \centering
            \includegraphics[width=\linewidth]{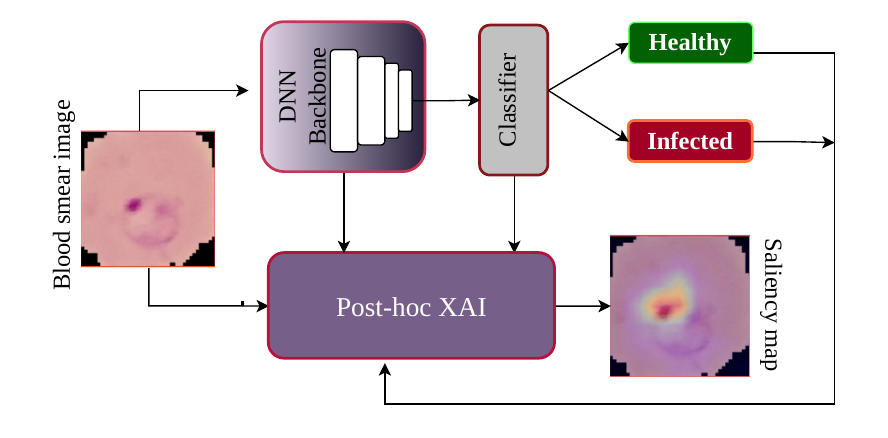}
            \caption{\textbf{Overview of the proposed framework.} A blood smear image is processed by a DNN backbone for binary classification, followed by a post-hoc XAI method that generates saliency maps highlighting the image regions most influential to the prediction.}
            \label{fig1}
        \end{figure}
        
        The proposed framework combines a deep learning classification pipeline with post-hoc explainability (Fig.\,\ref{fig1}). An input blood smear image is processed by a deep neural network (DNN) backbone that extracts high-level visual representations, which are subsequently passed to a fully connected classification head with a softmax output layer to generate a binary prediction: healthy or infected. Post-hoc XAI methods are then applied to the trained model by leveraging intermediate feature representations and gradient information to produce saliency maps highlighting the image regions that most strongly influence the model's decision.
        
  \subsection{Statistical analysis}
      \label{sec:statistical}
        Performance differences across classifiers are evaluated using the two-stage non-parametric protocol proposed by \cite{demvsar2006statistical}. 
        Let $r_{ij}$ denote the rank of classifier $M_i$ ($i=1,\dots,k$) on fold $j$ ($j=1,\dots,n$). The \emph{Friedman omnibus test}~\cite{friedman1937use} is first used to determine whether statistically significant performance differences exist among the classifiers:
        \begin{equation}
            \chi^2_F = \frac{12n}{k(k+1)} \sum_{i=1}^{k}\!\left(\bar{r}_i - \tfrac{k+1}{2}\right)^2,
            \label{eq:friedman}
        \end{equation}
        where $\bar{r}_i = n^{-1}\!\sum_j r_{ij}$.
        
        Under the null hypothesis of equivalent classifier performance, $\chi^2_F \sim \chi^2_{k-1}$. When the omnibus test yields $p < \alpha$, pairwise \emph{Wilcoxon signed-rank tests}~\cite{wilcoxon1945individual} are subsequently performed on the fold-level performance differences $d_j^{(i,i')} = f_{ij} - f_{i'j}$. To control the family-wise error rate, the Bonferroni correction is applied, resulting in a per-comparison significance threshold of $\alpha' = \alpha\big/\!\binom{k}{2}$~\cite{garcia2010advanced}.
        Two classifiers are considered statistically distinguishable if $p < \alpha'$. As a distribution-free framework, this protocol does not require normality assumptions and is therefore well suited for model comparisons under cross-validation settings.
    
\subsection{Explainable AI}
    Four attribution methods---Grad-CAM, Integrated Gradients, Score-CAM, and SHAP---are employed to provide both gradient-based and gradient-free explanations of how spatial image features contribute to classification outcomes. 

    \paragraph{\textit{Grad-CAM}.} Grad-CAM~\cite{selvaraju2017grad} produces a coarse localization map by computing the gradient of the target class score $y^c$ with respect to the $k$-th feature map $A^k$ of a chosen convolutional layer. These gradients are globally average-pooled across spatial locations to yield channel-wise importance weights:
    \begin{equation}
        \label{eqn2}
        \alpha_k^c = \frac{1}{Z} \sum_{i,j} \frac{\partial y^c}{\partial A^k_{ij}},
    \end{equation}
    where $Z$ denotes the number of spatial locations in $A^k$. The resulting saliency map $L^c = \mathrm{ReLU}\!\left(\sum_k \alpha_k^c A^k\right)$ retains only activations that positively influence the predicted class. However, evaluating gradients at a single point makes Grad-CAM susceptible to saturation and local noise~\cite{wang2020score}. \vspace{0.1cm}

    \paragraph{\textit{Integrated Gradients}.} Itgrd-Grad~\cite{sundararajan2017axiomatic} addresses gradient saturation by accumulating attribution signals along a straight path from a reference baseline $\bar{x}$ to the input $x$:
    \begin{equation}
        \label{eqn3}
        \phi_i(x) = (x_i - \bar{x}_i)\int_{0}^{1}
        \frac{\partial F\!\left(\bar{x} + \alpha(x - \bar{x})\right)}
        {\partial x_i}\,d\alpha,
    \end{equation}
    where $F$ is the model output for the target class and $\alpha \in [0,1]$ parameterizes the interpolation. This formulation satisfies the \emph{completeness} axiom~\cite{sundararajan2017axiomatic}, ensuring that summed attributions equal the difference in model output between input and baseline: $\sum_i \phi_i(x) = F(x) - F(\bar{x})$. 

    \paragraph{\textit{Score-CAM}.} Score-CAM~\cite{wang2020score} eliminates gradient dependence entirely by using forward-pass confidence scores as channel weights. Each activation map $A^k$ is upsampled to input resolution, normalised to $[0,1]$, and applied as a mask $\tilde{A}^k$ to the input. The weight for channel $k$ is the model's target-class score under that mask:
    \begin{equation}
        \label{eqn4}
        w_k^c = F^c\!\left(\tilde{A}^k \circ x\right),
    \end{equation}
    where $\circ$ denotes element-wise multiplication. The final attribution map is defined as $S^c = \mathrm{ReLU}\!\left(\sum_k \hat{w}_k^c A^k\right)$ using softmax-normalised weights $\hat{w}_k^c$, which improves robustness to gradient noise and numerical instability. 

    \paragraph{\textit{SHAP}.} SHapley Additive exPlanations (SHAP)~\cite{lundberg2017unified} is grounded in cooperative game theory. Each input feature (pixel) is assigned a Shapley value $\phi_i$, representing its average marginal contribution to the model output across all possible feature subsets. The gradient-based SHAP variant is used, combining sampling over a background distribution with expected gradients to approximate Shapley values efficiently for deep neural networks:
    \begin{equation}
    \label{eqn5}
        \phi_i = \mathbb{E}_{\bar{x}\sim\mathcal{D}}\!\left[
        (x_i - \bar{x}_i)\int_{0}^{1}
        \frac{\partial F\!\left(\bar{x} + \alpha(x - \bar{x})\right)}
        {\partial x_i}\,d\alpha\right],
    \end{equation}
    where the expectation is taken over background samples $\bar{x}$ drawn from a reference distribution $\mathcal{D}$.

\section{Implementations and Results}
    \subsection{Data}
\begin{figure}[H]
            \centering
            \setlength{\tabcolsep}{2pt}       
            
            \begin{subfigure}[t]{0.29\linewidth}
                \includegraphics[width=\linewidth]{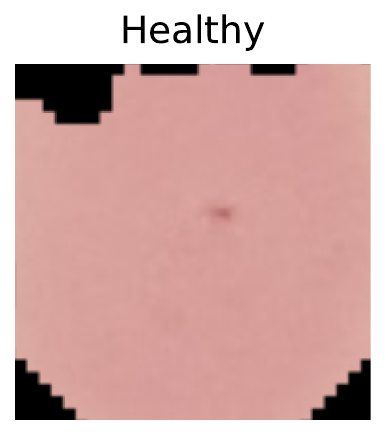}
            \end{subfigure}
            \begin{subfigure}[t]{0.29\linewidth}
                \includegraphics[width=\linewidth]{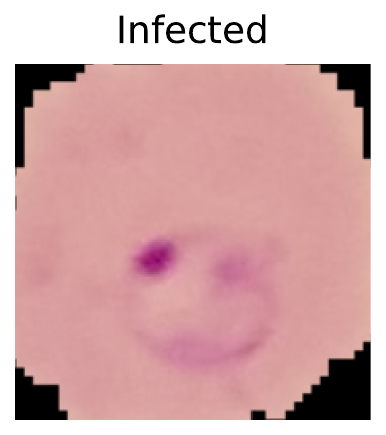}
            \end{subfigure}
            \begin{subfigure}[t]{0.38\linewidth}
                \includegraphics[width=\linewidth]{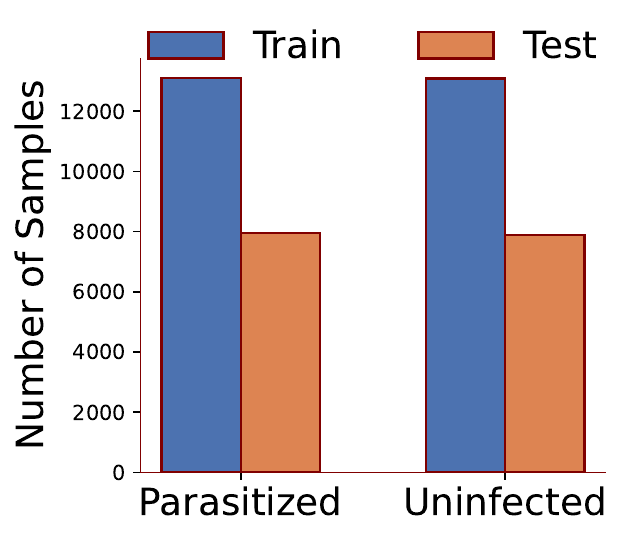}
            \end{subfigure}
            \caption{\textbf{Overview of the NLM-Malaria dataset.} Representative examples of uninfected (left) and infected (center) blood smear cells, alongside the training and test set class distributions (right).}
            \label{fig:dataset}
        \end{figure}

        The NLM-Malaria dataset~\cite{rajaraman2018pre} is used (Fig.\,\ref{fig:dataset}), a publicly available collection of thin blood smear cell images split into training and test sets. 

        The training set contains $27,560$ images ($13,780$ infected and $13,780$ uninfected), while the test set contains $15,832$ images ($7,952$ infected and $7,880$ uninfected), corresponding to an approximately 63/37 split.
        All images were resized to $256\times 256$ and center-cropped to $224 \times 224$, followed by normalization using ImageNet mean and standard deviation. Standard data augmentation techniques---random rotation, horizontal flipping, translation, and cropping---are applied during training\footnote{Code available \href{https://anonymous.4open.science/r/efficient-malaria-detection-xai-CB0E/README.md}{https://github.com/efficient-malaria-xai}}. No synthetic samples are generated; augmentation is performed on the fly during training.
                
    \subsection{Model Hyperparameters and Training}
        All models, including ResNet, Vision Transformer (ViT), EfficientNet, and MobileNet are sourced from Torchvision and initialized with ImageNet pre-trained weights. Each model is fine-tuned end-to-end after replacing the classifier head to match the binary target task. For each fold, the checkpoint achieving the best validation performance is retained for evaluation. Shared hyperparameters are reported in Table~\ref{hyperparameters}, with the exception of ViT, which uses a different learning rate.
        Experiments are conducted on an NVIDIA GeForce GTX 1060 (6GB VRAM) using PyTorch with CUDA. Due to the GPU memory constraints, a batch size of 4 is used.
        \begin{table}[H]
         \centering
            \caption{Training hyperparameters used across all experiments.}
            \resizebox{\columnwidth}{!}{
                \small
                \begin{tabular}{ll}
                    \toprule
                    \textbf{Hyperparameter} & \textbf{Value} \\
                    \midrule
                    Epochs        & 100 with early stopping on 5 epochs (val. acc.) \\
                    Learning rate           & $1\times10^{-3}$ (CNN), $1\times10^{-4}$ (ViT) \\
                    Number of folds         & 5 (Stratified cross-validation) \\
                    Loss function           & Cross entropy \\
                    \bottomrule
                \end{tabular}
            }
            \label{hyperparameters}
        \end{table}        

    \subsection{Model Evaluation}
        \paragraph{Clinical relevance metrics.} Sensitivity and specificity are fundamental measures of diagnostic performance, capturing a model’s ability to correctly identify positive (diseased) and negative (non-diseased) cases, respectively. Positive predictive value (PPV) and negative predictive value (NPV) are additionally reported, as they provide complementary clinically relevant information about test performance. 
        All metrics are summarized in Table~\ref{specificity_sensitivity} to ensure a clinically interpretable evaluation of the proposed models.    
        \begin{table}[H]
            \centering
            \caption{Clinical relevance metrics of the models}
            \resizebox{\columnwidth}{!}{
            \begin{tabular}{lcccc}
            \toprule
            \textbf{Model} & \textbf{Sensitivity} & \textbf{Specificity}  & PPV & NPV \\
            \midrule
            ResNet-18 & 0.984 & 0.955 & 0.956 & 0.984\\    
            MobileNetV3 & 0.976 & 0.963 & 0.964 & 0.976\\
            EfficientNetB0 & 0.971 & 0.965 & 0.965 & 0.972 \\
            ViT-B/16 & 0.970 & 0.948 & 0.949 & 0.969\\
            \bottomrule
            \end{tabular}
            }
            \label{specificity_sensitivity}
        \end{table}
        
\begin{figure}[H]
            \centering
            \includegraphics[width=\linewidth]{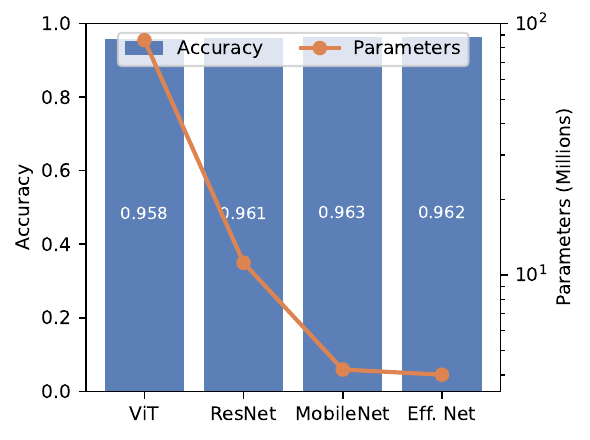}
            \caption{\textbf{Accuracy and parameter count across architectures}. Bars (y-axis) show mean cross-validated accuracy; the overlaid line (x-axis) indicates the number of trainable parameters.}
            \label{acc.vs.params}
        \end{figure}

        \paragraph{Predictive performance.} Model performance is evaluated using 5-fold cross-validation to ensure robust and unbiased estimation. For each fold, the training is performed on the training subset and evaluated on the held-out validation split, with metrics aggregated across folds.   
        Accuracy, F1-score, and AUC are reported as means $\pm$ standard deviation across the five folds. The area under the receiver curve (AUC) is computed from predicted class probabilities using a one-vs-rest strategy with macro averaging. Given the limited number of folds, $95\%$ confidence intervals are estimated using Student's \textit{t}-distribution to account for uncertainty in the variance estimate. Figure~\ref{acc.vs.params} shows accuracy against  as a function of model size (parameter count). Detailed training and validation results, including accuracy, precision, recall, F1-score, and AUC, are provided in Appendix \ref{appendix}.

        \paragraph{Computational costs.} Beyond the accuracy–parameter trade-off, model efficiency is evaluated using floating-point operations (FLOPs), per-image inference latency (mean $\pm$ standard deviation over 500 runs), and peak CPU memory consumption during end-to-end inference for a single image. All measurements are conducted on a system equipped with a 12-core Intel processor and 15.44 GB of RAM. The results are summarized in Table~\ref{efficiency}.
        \begin{table}[H]
            \centering
            \caption{Model efficiency under CPU-based inference, reported in terms of parameter count (M), FLOPs (G), per-image latency (ms), and peak memory usage (MB).}
            \resizebox{\columnwidth}{!}{
            \begin{tabular}{lcccc}
            \toprule
            \textbf{Model} & \textbf{Params} & \textbf{FLOPs} & \textbf{Latency} & \textbf{Peak Memory} \\
            \midrule
            ResNet-18 & 11.18 & 1.82 & 20.92 $\pm$ 1.78 & 791.74 \\    
            MobileNetV3 & 04.20 & 0.22 &14.37 $\pm$ 2.31 & 557.04\\
            EfficientNetB0 & 04.01 &0.40 &37.19 $\pm$ 4.31 & 493.25\\
            ViT-B/16 & 85.80 & 16.8 & 293.8 $\pm$ 49.18 & 843.59\\
            \bottomrule
            \end{tabular}
            }
            \label{efficiency}
        \end{table}
        
        \paragraph{Ensembling results.} Individual model and ensemble classification performance are reported in Table~\ref{ensembling}. The ensemble is constructed by aggregating raw logits across models before the softmax layer and deriving the final prediction. 
        Logit-level aggregation is preferred over majority voting, as the latter can provide uninformative ties in the case of an even number of models---resulting in a $50/50$ split under equal disagreement---whereas logit averaging preserves the full confidence distribution of each model and yields a more calibrated combined prediction. 
        Among individual models, ViT performs below the CNN-based architectures, registering the lowest accuracy ($0.958\%$), F1-score ($0.959$), and AUC ($0.990$). 
        This performance gap is consistent with the reported sample inefficiency of vision transformers relative to CNNs \cite{lu2022bridging}. The higher parameter count and reliance on large-scale pretraining reduce the effectiveness of ViT on relatively small, domain-specific datasets such as malaria cell images, where CNN-based architectures remain more competitive.
        \begin{table}[H]
            \centering
            \caption{Classification performance of individual models and the ensemble. Means $\pm$ std are reported across five cross-validation folds.}
            \resizebox{\columnwidth}{!}{
            \begin{tabular}{lccc}
            \toprule
            Model & Accuracy & Macro-F1 & Macro-AUC \\
            \midrule
            ResNet-18 & 0.961 $\pm$ 0.003 & 0.961 $\pm$ 0.003 & 0.991 $\pm$ 0.002 \\
            MobileNet-v3-l & 0.963 $\pm$ 0.005 & 0.963 $\pm$ 0.005 & 0.992 $\pm$ 0.002 \\
            EfficientNet-b0 & 0.962 $\pm$ 0.003 & 0.962 $\pm$ 0.003 & 0.992 $\pm$ 0.002 \\
            ViT\_b\_16 & 0.958 $\pm$ 0.004 & 0.958 $\pm$ 0.004 & 0.989 $\pm$ 0.002 \\
            \textbf{Ensemble} & \textbf{0.972} &\textbf{0.972}&\textbf{0.996 }\\            
            \bottomrule
            \end{tabular}
            }
            \label{ensembling}
        \end{table}

    \paragraph{Statistical analysis.} For statistical comparison across models, the Friedman test was applied as described in Sec.~\ref{sec:statistical}. Across all reported metrics, the test yields $\chi^2_F = 2.5000$ and $\text{p-value} = 0.4717$ at a significance level $\alpha = 0.01$, indicating no statistically significant performance differences among the four models. This result indicates that, despite substantial variation in model size, all architectures achieve comparable diagnostic performance, supporting the relevance of efficient-by-design models in resource-constrained settings.        

    \subsection{Robustness Analysis}
        Model robustness is evaluated under three types of perturbations~\cite{hendrycks2019benchmarking}, reflecting the noisy acquisition conditions commonly encountered in low-resource clinical settings.  
        \begin{figure}[H]
            \centering            \includegraphics[width=\linewidth]{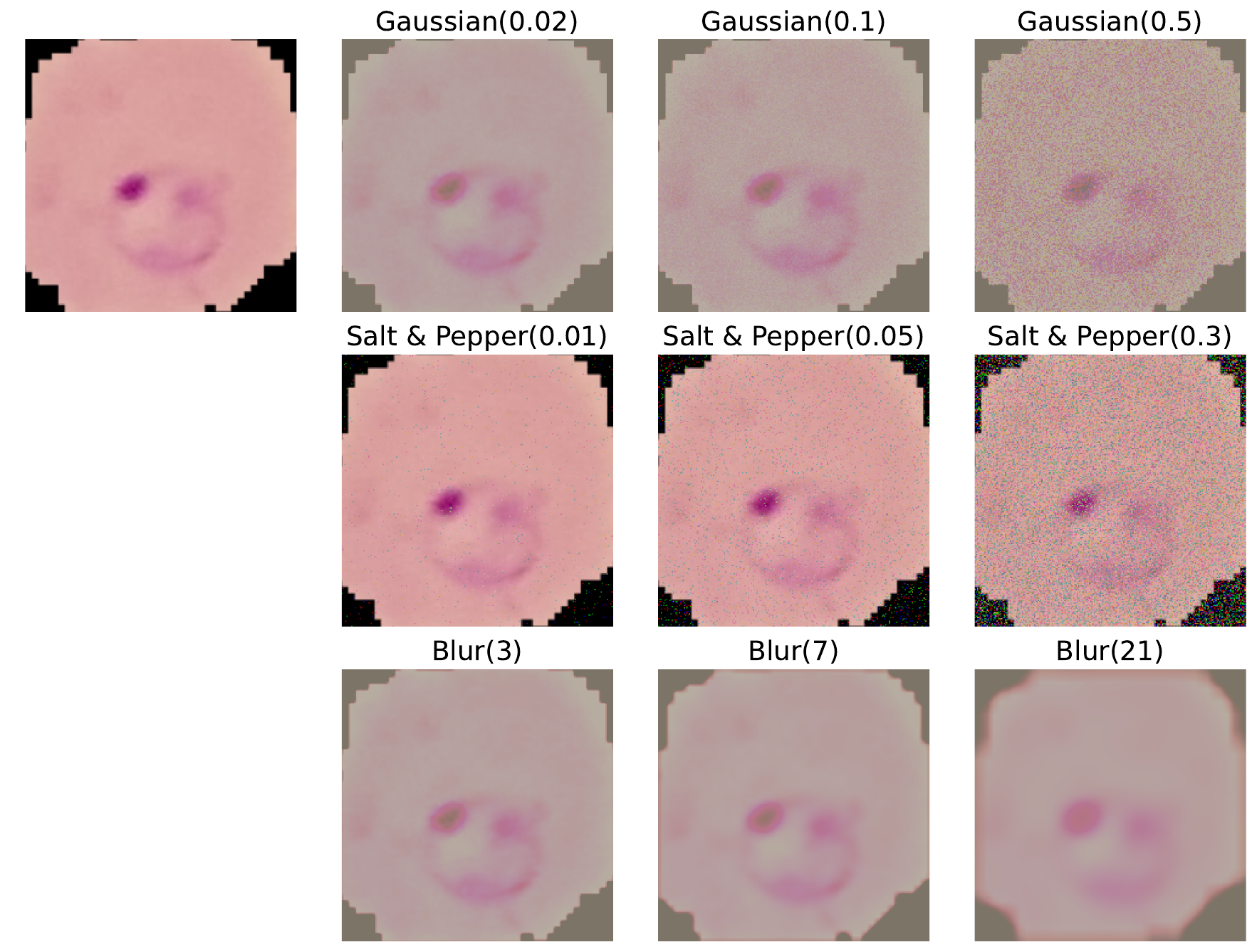}
            \caption{Example of an infected blood smear cell images across different noise types and severity levels.}
            \label{corruption}
        \end{figure}          
        Test images are systematically corrupted at increasing severity levels using Gaussian noise ($\sigma \in \{0.02, 0.05, 0.10, 0.20, 0.35, 0.50\}$), salt-and-pepper noise (corruption rate $\in \{0.01, 0.03, 0.05, 0.10, 0.20, 0.30\}$), and Gaussian blur (kernel size $\in \{3, 5, 7, 11, 15, 21\}$ pixel). These perturbations simulate common microscopy artifacts, including sensor noise, transmission errors, and optical defocus. Representative corrupted images at mild, moderate, and severe severity levels are shown in Figure~\ref{corruption}.
        For each corruption level, model performance is quantitatively assessed using mean predicted confidence and mean accuracy computed over 50 randomly sampled, correctly classified images per class. Predicted confidence is defined as the softmax probability assigned to the ground-truth class. Together, these metrics provide a comprehensive assessment of how reliably each model maintains its predictions under increasing levels of perturbation.
        \begin{figure}[H]
            \centering
            \setlength{\tabcolsep}{1pt}     
            \scriptsize 
            \begin{tabular}{c c c c}            
                & \makebox[0.28\columnwidth]{\hspace{0.25cm}Gaussian} & \makebox[0.28\columnwidth]{\hspace{0.25cm} Salt \& Pepper}
                 & \makebox[0.28\columnwidth]{\hspace{0.25cm}Blur} \\
                \rotatebox{90}{\hspace{1cm} ResNet}&
                \includegraphics[width=0.3\columnwidth]{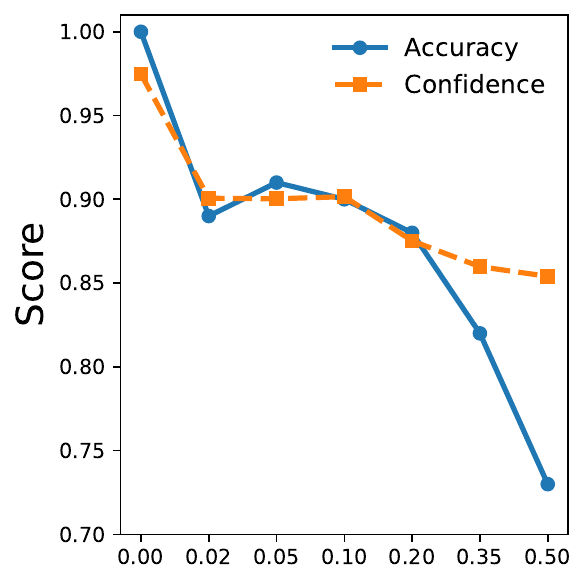} &
                \includegraphics[width=0.3\columnwidth]{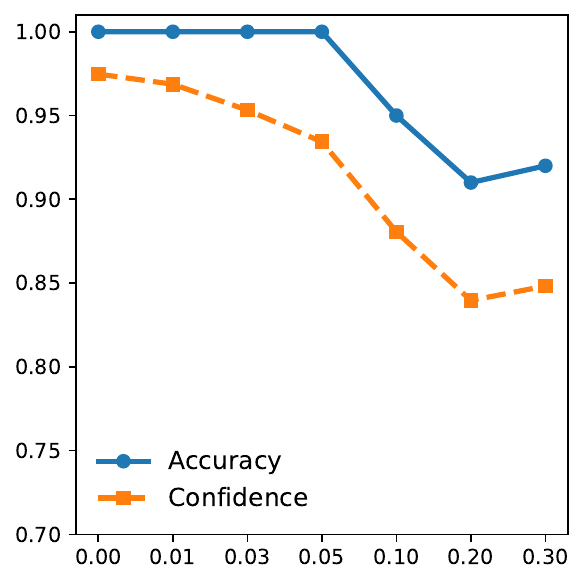} &
                \includegraphics[width=0.3\columnwidth]{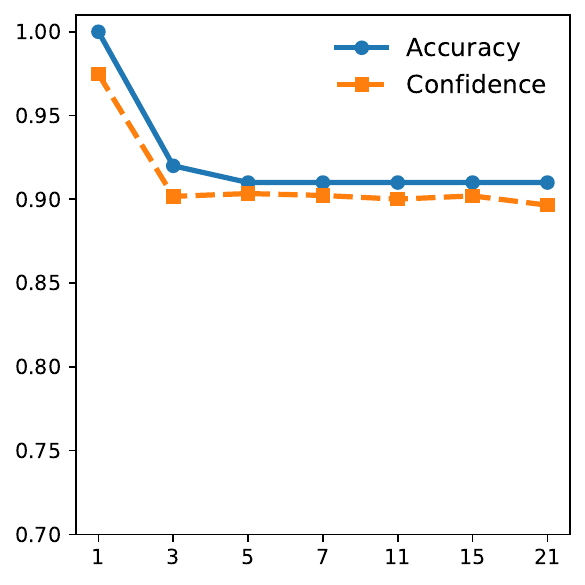} \\
            
                \rotatebox{90}{\hspace{0.9cm} MobileNet} &
                \includegraphics[width=0.3\columnwidth]{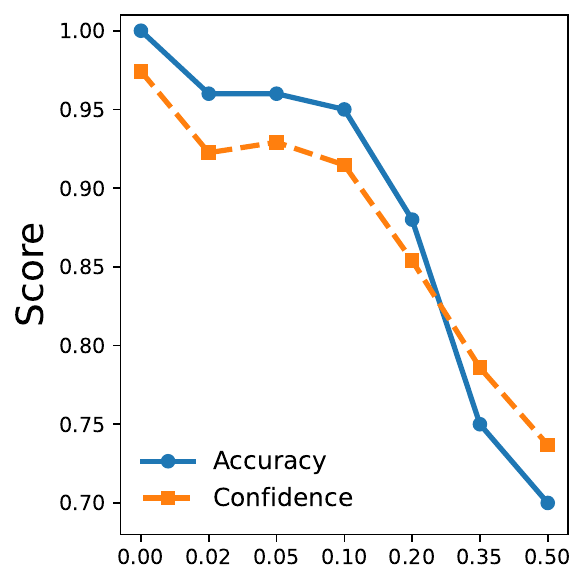} &
                \includegraphics[width=0.3\columnwidth]{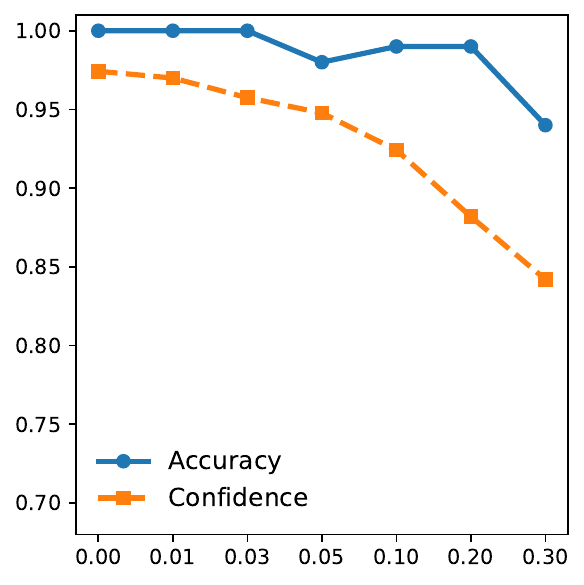} &
                \includegraphics[width=0.3\columnwidth]{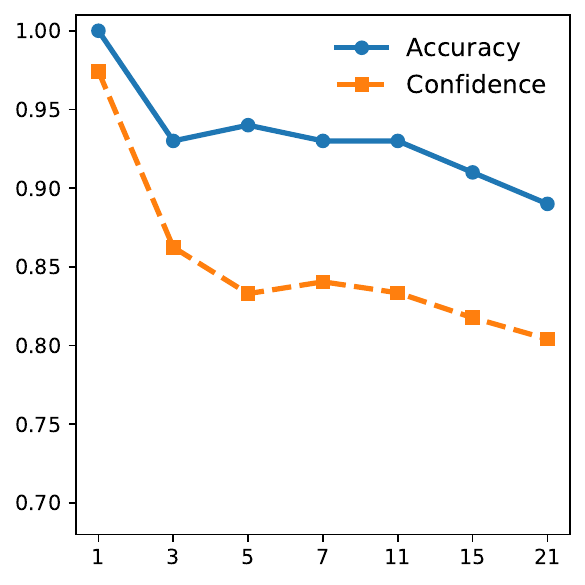} \\
                
                \rotatebox{90}{\hspace{0.8cm} EfficientNet} &
                \includegraphics[width=0.3\columnwidth]{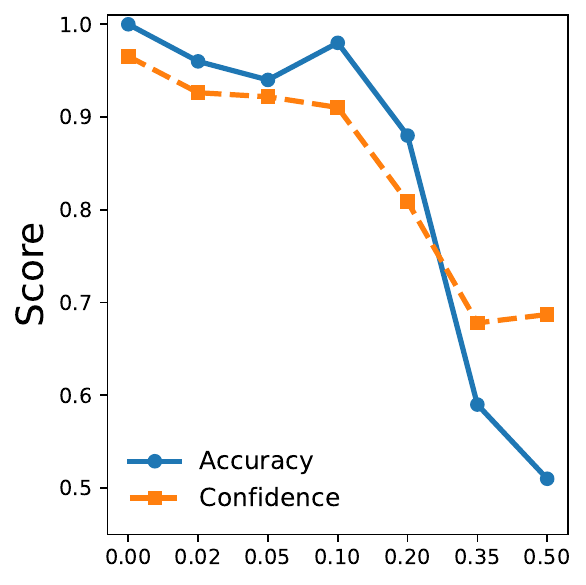} &
                \includegraphics[width=0.3\columnwidth]{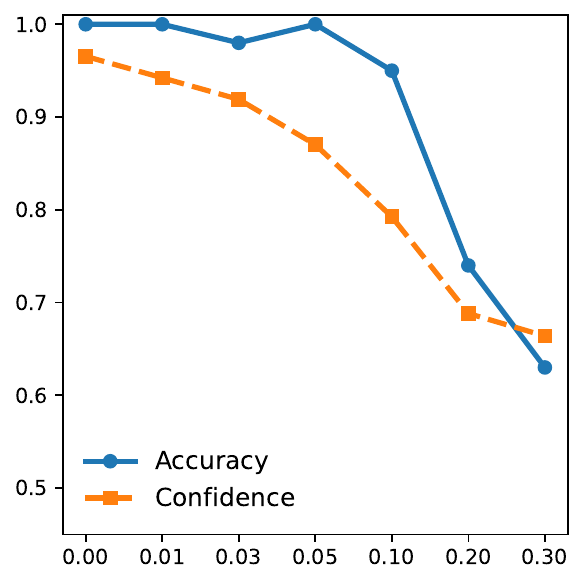} &
                \includegraphics[width=0.3\columnwidth]{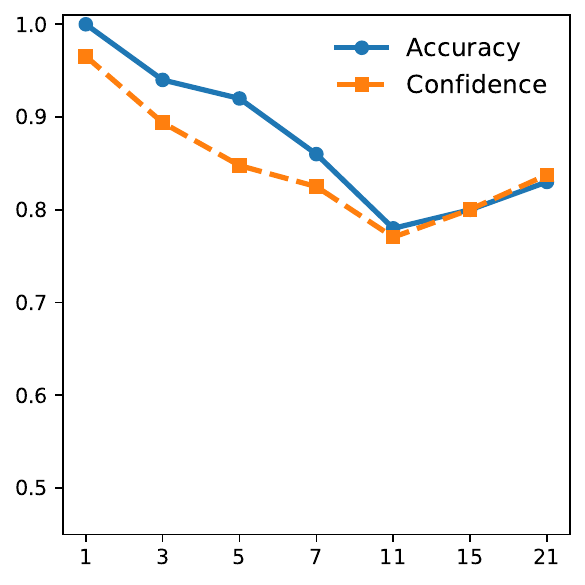} \\
                
                \rotatebox{90}{\hspace{1.2cm} ViT} &
                \includegraphics[width=0.3\columnwidth]{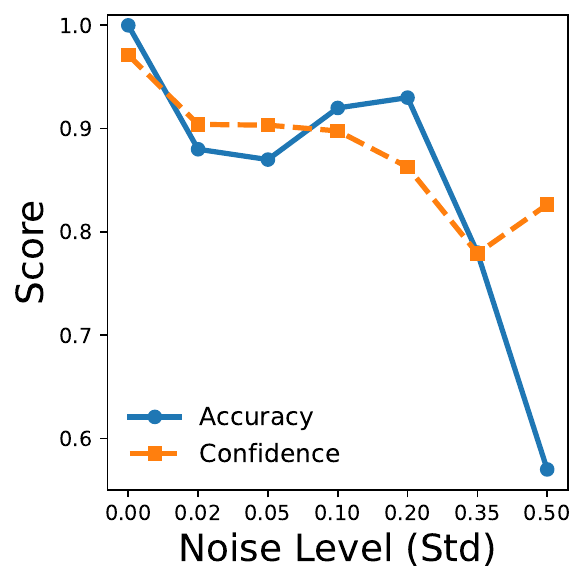} &
                \includegraphics[width=0.3\columnwidth]{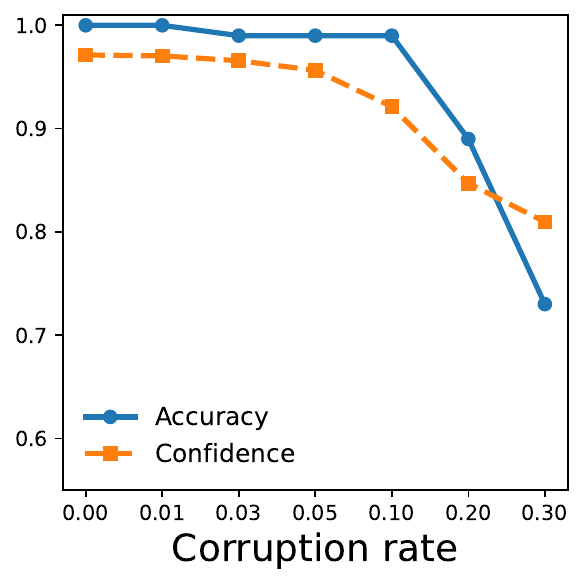} &
                \includegraphics[width=0.3\columnwidth]{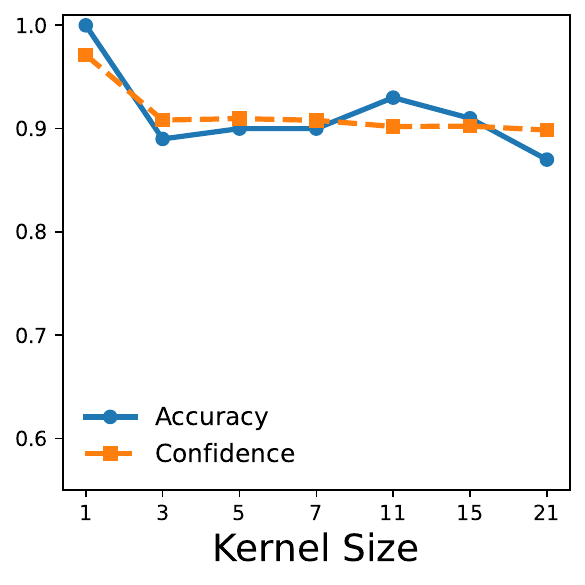} \\            
            \end{tabular}            
            \caption{Performance under different noise types across architectures. Rows correspond to models and columns to noise types.}
            \label{sensitivitymodels}
        \end{figure}   
        
        Model confidence and accuracy as a function of perturbation severity are illustrated in Figure~\ref{sensitivitymodels}. All models demonstrate strong robustness to salt-and-pepper noise. 
        ResNet-18 maintains 100\% robust accuracy through the first three corruption levels, although confidence declines to approximately 94\%, while MobileNet sustains perfect accuracy through the first two levels.
        Gaussian noise proves substantially more disruptive across all architectures, with MobileNetV3 achieving the strongest performance,  although accuracy still decreases to 96\%  at the lowest severity level.
        Gaussian blur presents a similar challenge, with EfficientNet-B0 achieving the highest accuracy of 94\% at the first severe corruption level. Complete numerical results are provided in Appendix~\ref{sensitivityAppendix}.
    
    \subsection{Qualitative explanation analysis}        
        All four post-hoc CNN-based attribution methods are implemented using standard Python libraries. SHAP attributions are computed using GradientExplainer from shap library, while the remaining methods are implemented using PyTorch primitives, including \texttt{register\_forward\_hook} and \texttt{register\_backward\_hook}. 

\begin{figure}[H]
            \centering
            \includegraphics[width=\linewidth]{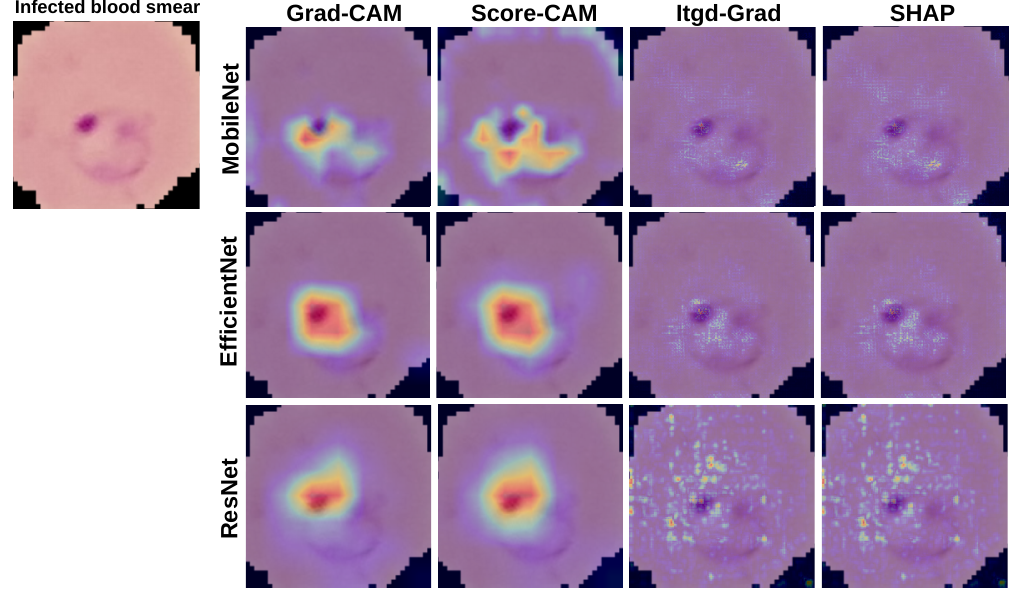}
            \caption{Saliency-based explanation maps across CNN architectures on an infected blood smear image. }
            \label{explanations}
        \end{figure}

        \paragraph{Qualitative visualization of CNN models.}
        Attribution maps produced by Grad-CAM, Score-CAM, Integrated Gradients (Itgd Grad), and SHAP for a representative infected cell across all CNN architectures are illustrated in Figure~\ref{explanations}. 
        Integrated Gradients and SHAP generate fine-grained, pixel-level explanations, while Grad-CAM and Score-CAM produce coarser region-level saliency maps.
        Across architectures, several consistent patterns emerge: EfficientNet-B0 and ResNet-18 produce similar Grad-CAM and Score-CAM maps, with attribution concentrated near the image center where the parasite is located. In contrast, Integrated Gradients and SHAP explanations for ResNet-18 exhibit significantly higher noise levels, with high-attribution pixels distributed broadly across the image rather than localized to the infected region.
        
        \paragraph{Consistency setup across explanation.} Spatial agreement between post-hoc explanation methods is analyzed to identify refions where attributions converge or diverge across image, including parasite boundaries and background regions. Two intra-family comparisons are considered: Grad-CAM vs.\ Score-CAM (coarse, region-level explanations) and Integrated Gradients vs.\ SHAP (fine-grained explanations). Given two saliency maps $S_A$ and $S_B$, each map is independently min-max normalized, and the absolute difference map is computed as $D(i,j) = |S_A(i,j) - S_B(i,j)|$ with values constrained to $[0,1]$. 
        To isolate statistically meaningful disagreement rather than numerical noise, the difference map $D$ is binarized using an adaptive threshold defined by
        $B(i,j) = \mathbb{1}[D_{\text{dis}}(i,j) > \mu + k\cdot\sigma]$ where $\mu$ and $\sigma$ denote the empirical mean and standard deviation of $D$, $k$ controls threshold strictness, and $\mathbb{1}[\cdot]$ is the indicator function.
        An agreement map is similarly defined to identify regions where both methods simultaneously assign high attribution. Rather than averaging the two maps---which may yield high values even when one method assigns strong importance---independent thresholds are applied to each map followed by a logical conjunction: $B_{\text{agr}}(i,j) = \mathbb{1}\bigl[S_A(i,j) > \mu_A + k\cdot\sigma_A\bigr] \;\wedge\; \mathbb{1}\bigl[S_B(i,j) > \mu_B + k\cdot\sigma_B\bigr]$, where $\mu_A, \sigma_A$ and $\mu_B, \sigma_B$ denote the mean and standard deviation of $S_A$ and $S_B$ respectively. Pixels flagged by $B_{\text{agr}}$ correspond to region independently identified as salient by both methods, thereby providing stronger evidence of genuinely important regions. The disagreement map $B_{\text{dis}}$ and agreement map $B_{\text{agr}}$ provide complementary views: the former highlights regions of conflict, whereas the latter identifies regions of converge.  A fixed $k=1.0$ is used throughout, with results for alternative values reported in Appendix~\ref{disagreement}.

\begin{figure}[H]
                \centering
                \includegraphics[width=\linewidth]{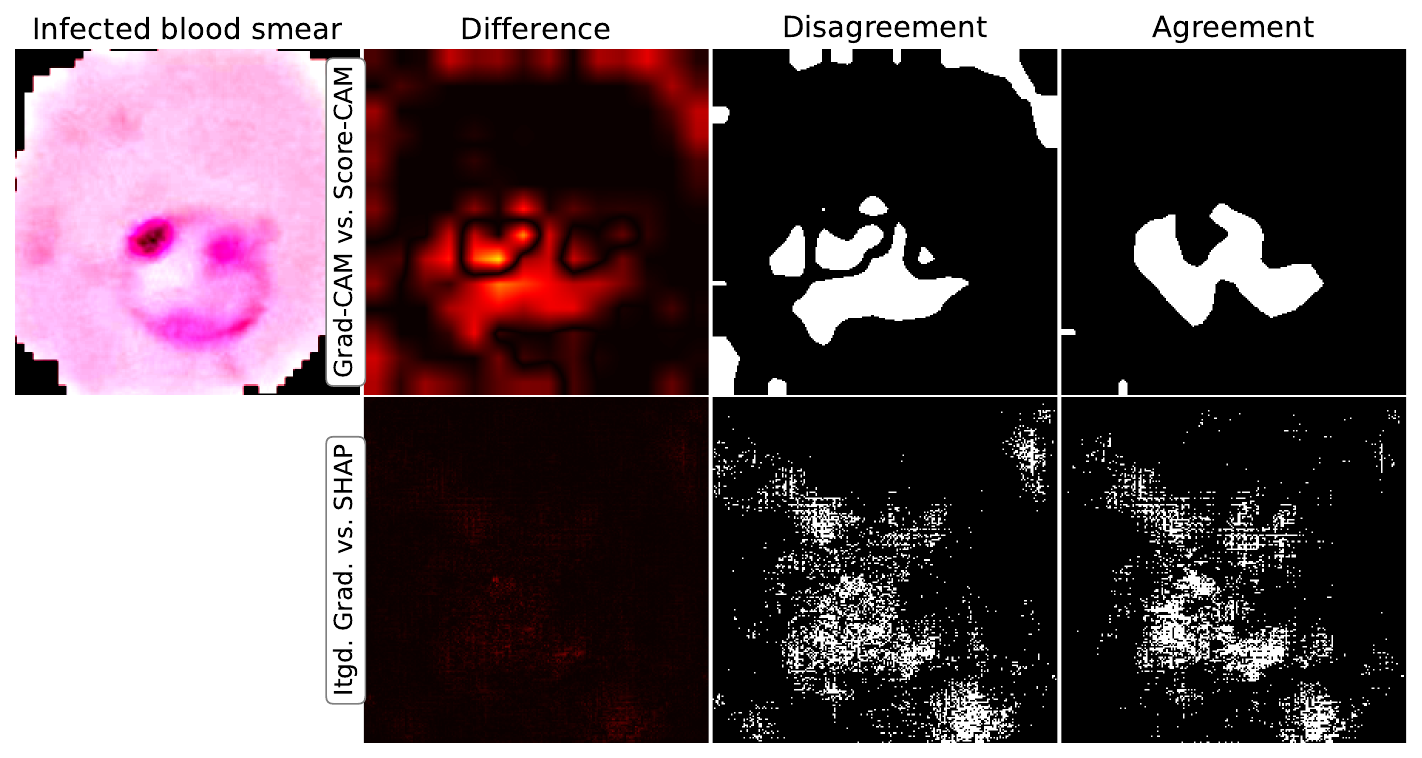}
            \caption{Disagreement maps on MobileNet.}
            \label{fig:disagreement}
        \end{figure}

\begin{figure}[H]
                \centering
                \begin{subfigure}{\linewidth}
                    \includegraphics[width=\linewidth]{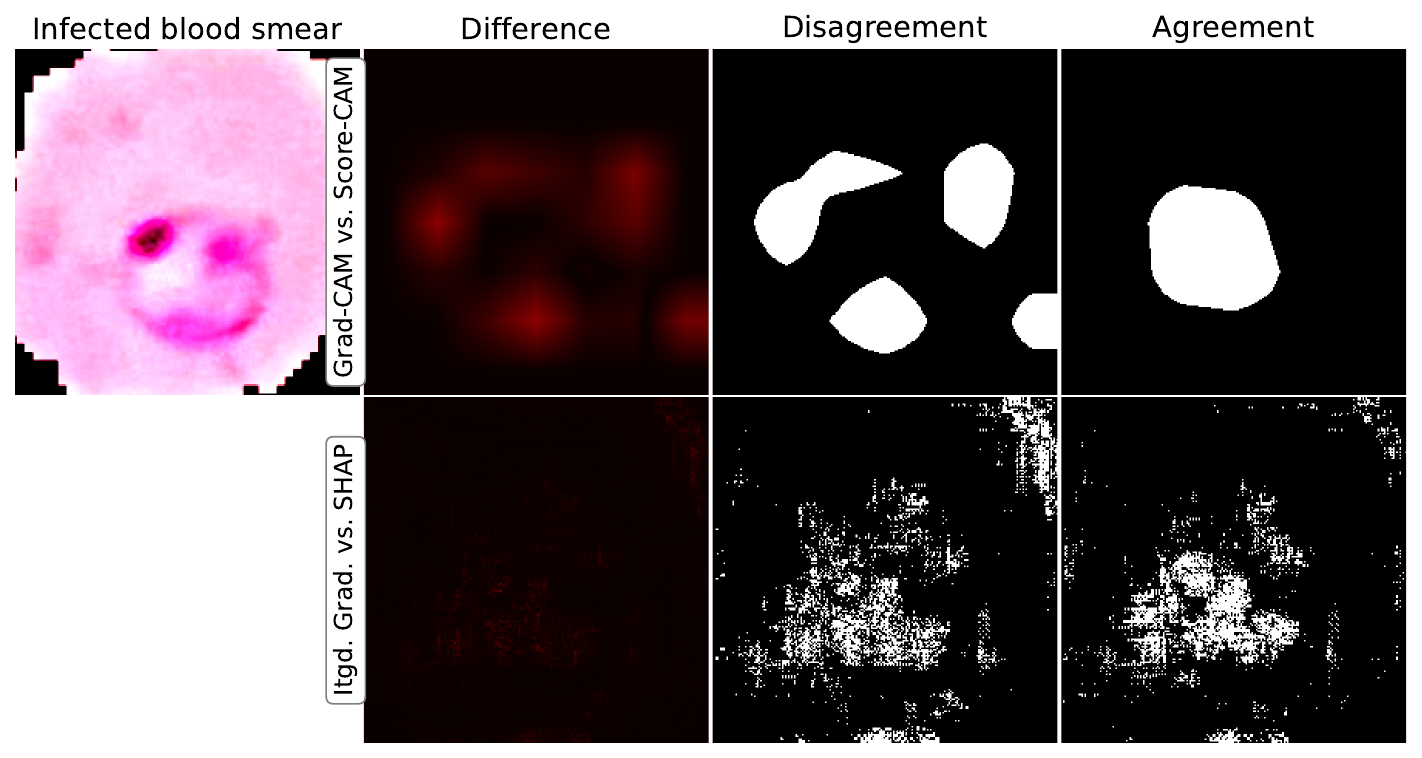}
                    \caption{EfficientNet} 
                \end{subfigure}

                \begin{subfigure}{\linewidth}
                    \centering
                    \includegraphics[width=\linewidth]{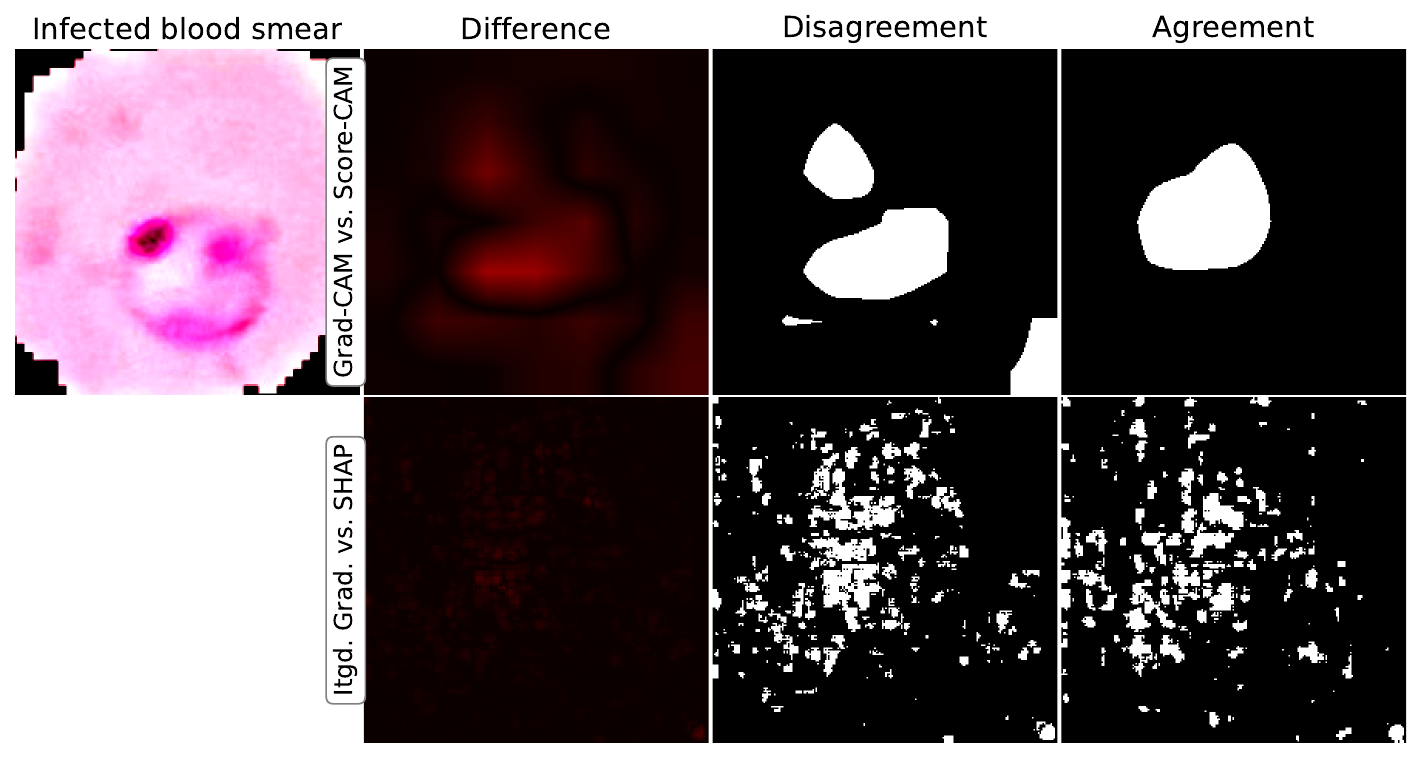}
                    \caption{ResNet-18} 
                    \label{dissagreerenet}
                \end{subfigure}
                \caption{Disagreement maps on EfficientNet and ResNet.}
                \label{fig:disagreement_all}
        \end{figure}

        \paragraph{Pairwise Explanation Consistency.} Figure~\ref{fig:disagreement} and~\ref{fig:disagreement_all} reveals several notable patterns. 
        Despite their shared region-level characteristics (Fig.~\ref{explanations}), Grad-CAM and Score-CAM do not consistently agree: their difference maps exhibit large, spatially coherent regions, indicating structured disagreement concentrated in anatomically meaningful areas rather than random noise. Similar patterns are observed in the corresponding agreement maps, is complement the disagreement analysis.
        In contrast, both disagreement and agreement between Integrated Gradients and SHAP manifests as diffuse, unstructured patterns similar to salt-and-pepper noise,making spatial interpretation considerably more diffucult. At the architectural level, the agreement and disagreement maps exhibit distinct spatial signatures. For EfficientNet-B0, disagreement between Grad-CAM and Score-CAM is concentrated near the outer parasite boundary whereas agreement is localized to the central bounded region. For ResNet-18, disagreement appears primarily within the parasite region itself, while agreement partially overlaps the chromatin dot, an important structure for identifying stage of parasite in trophozoite stage of the parasite. MobileNetV3 exhibits finer-grained agreement and disagreement patterns along detailed internal parasite boundaries and structures.
        
\begin{figure}[H]
            \centering
                \centering
                \includegraphics[width=\linewidth]{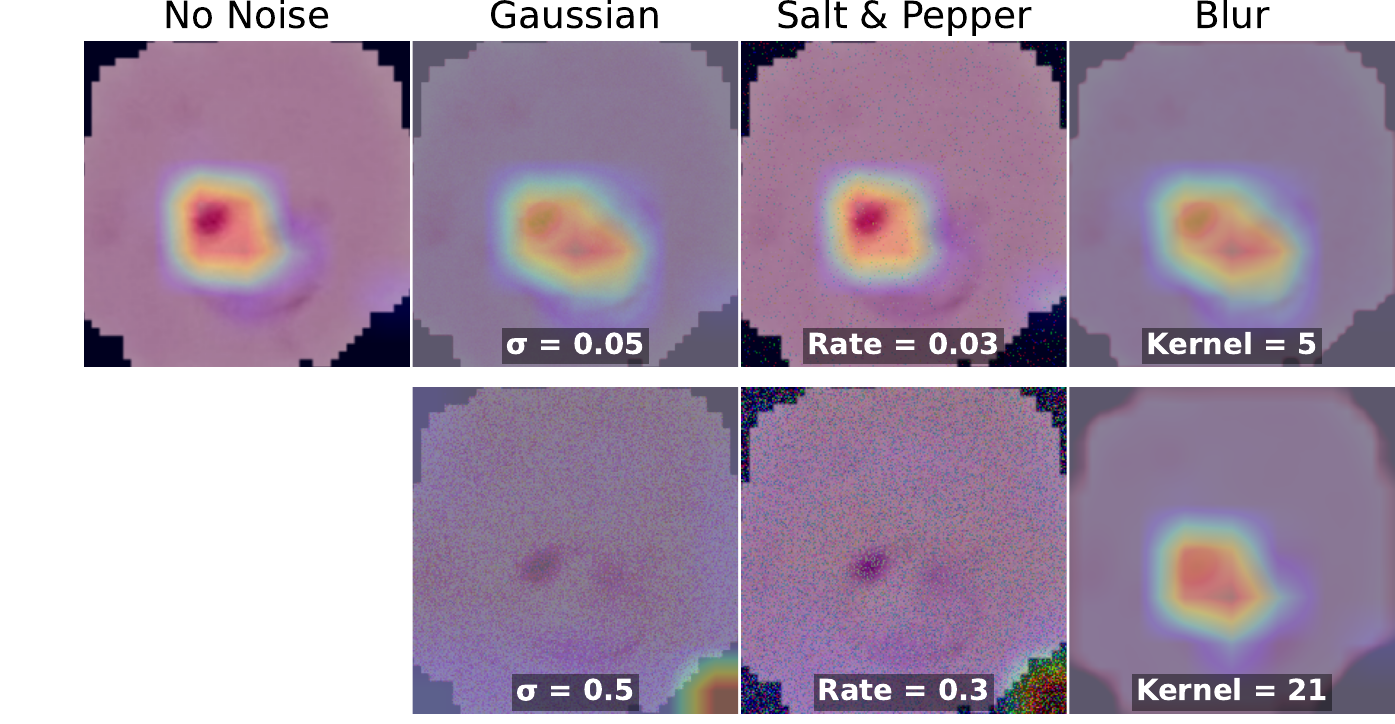}
            \caption{\textbf{Grad-CAM attribution maps for EfficientNet under increasing levels of perturbation}. The first row represents the mild level, while the second row represents the severe level under various perturbations (Gaussian, Salt \& Pepper, and Blur).}
            \label{corruptionxai}
        \end{figure}

         \paragraph{Consistency analysis under perturbations.} The stability of salient regions under input corruption is evaluated using the \emph{Spearman rank correlation} $\rho$ between saliency maps generated from clean and corrupted inputs. Figure~\ref{corruptionxai} provides a qualitative illustration using Grad-CAM on EfficientNet---the architecture exhibiting the lowest overall $\rho$ across corruption types---under three severity levels: no corruption, mild corruption, and severe corruption. The figure (second row) indicates that the model remains comparatively robust under severe blur perturbations, whereas severe Gaussian noise and salt \& pepper corruption substantially alter the explanation maps, leading to pronounced shifts in the salient regions. 

\begin{figure}[H]
            \centering
            \includegraphics[width=\linewidth]{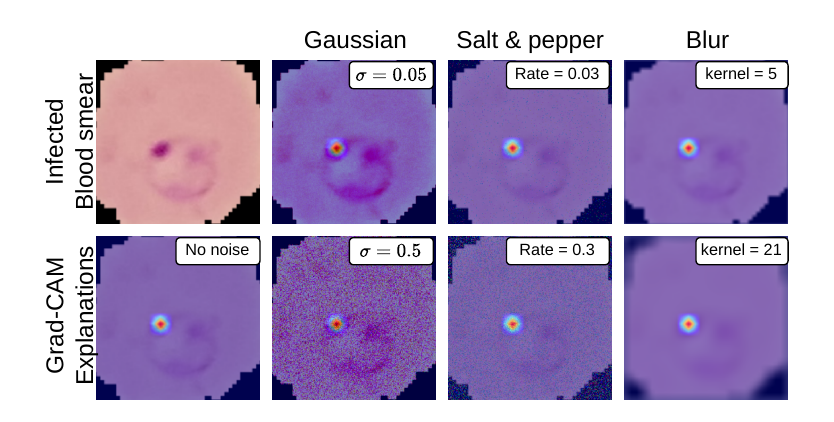}
            \caption{ViT explanation using Grad-CAM.}
            \label{vit_grad}
        \end{figure}

        \paragraph{Qualitative explanations for ViT.} 
        On the Vision Transformer model, Grad-CAM is adapted to analyze the spatial regions contributing to predictions (Fig.~\ref{vit_grad}). 
        The model consistently focuses on a small, highly localized region corresponding to the \textit{chromatin dot} of the ring-stage \textit{Plasmodium} parasite. 
        The resulting attention maps are sharply concentrated, with minimal activation across the surrounding erythrocyte cytoplasm, indicating that the Vision Transformer captures highly discriminative fine-grained morphological features. This behavior contrasts with typical convolutional architectures, which generally produce more spatially diffuse activation patterns. Notably, this localized focus remains stable even under input perturbations such as Gaussian corruption.

    \subsection{Quantitative explanation analysis}
\begin{figure}[H]
            \centering
            \setlength{\tabcolsep}{2pt}     
            \scriptsize
            \begin{tabular}{c c c c}   
                & \multicolumn{1}{c}{ResNet} 
                & \multicolumn{1}{c}{MobileNet}
                & \multicolumn{1}{c}{EfficientNet} \\
                
                \rotatebox{90}{\hspace{0.7cm} Insertion Score} &
                \includegraphics[width=0.3\columnwidth]{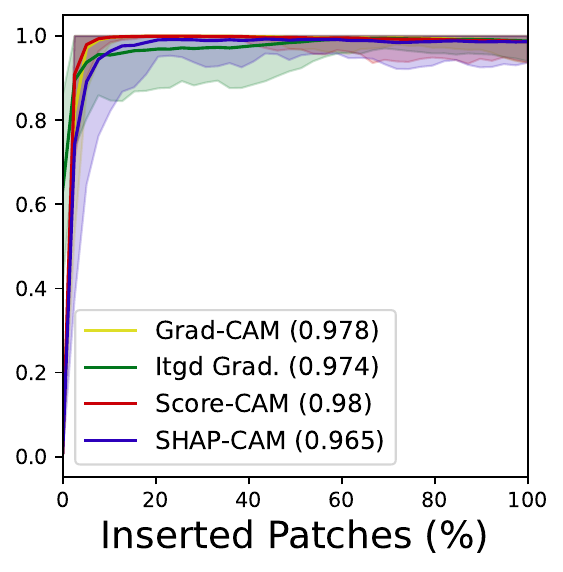} &
                \includegraphics[width=0.3\columnwidth]{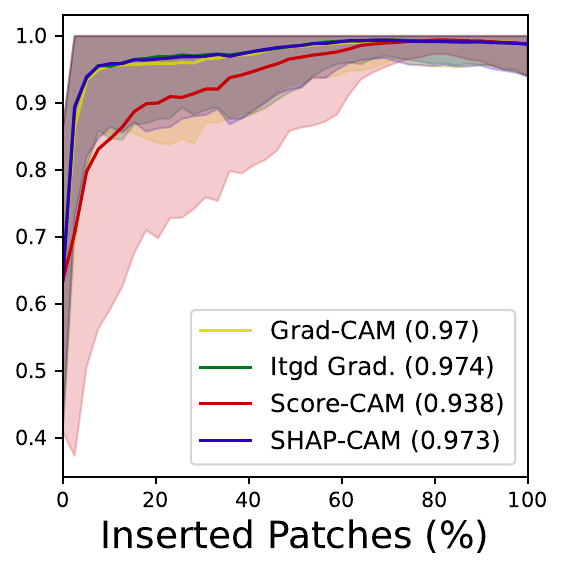} &
                \includegraphics[width=0.3\columnwidth]{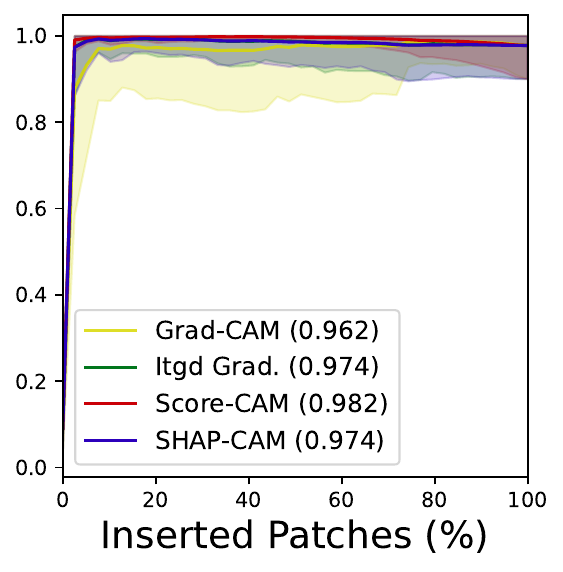} \\

                \rotatebox{90}{\hspace{0.7cm} Deletion Score} &
                \includegraphics[width=0.3\columnwidth]{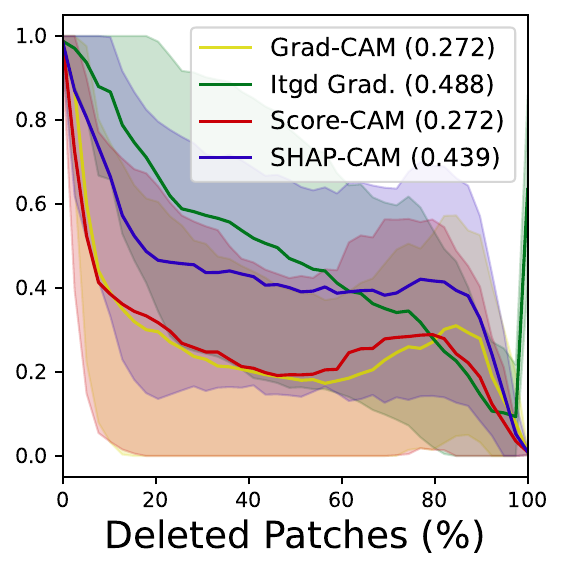} &
                \includegraphics[width=0.3\columnwidth]{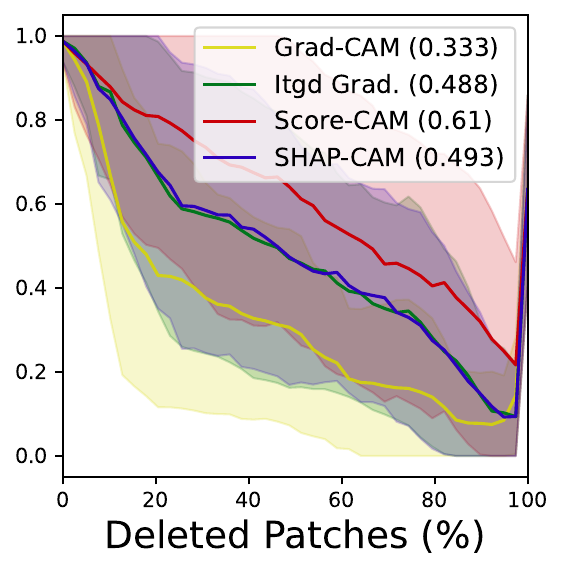} &
                \includegraphics[width=0.3\columnwidth]{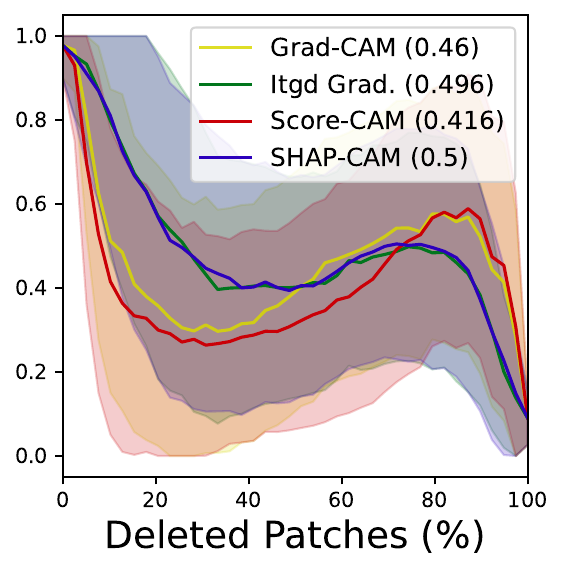} \\
                  
                \end{tabular}            
                \caption{\textbf{Insertion and Deletion curves across all CNN architectures.} The first row represents insertion curves, while the second row represents deletion curves.}
                \label{insertioncurves} 
            \end{figure}

        To quantitatively evaluate explanation quality, \emph{insertion} and \emph{deletion}~\cite{gomez2022metrics} are adopted as a complementary quantitative metrics. Insertion progressively reveals image regions in descending order of attributed importance and measures the corresponding increase in predicted class confidence, while deletion progressively occludes regions in the same order and measures the resulting confidence decrease (Fig.~\ref{insertioncurves}). Rather than operating at the pixel level, Each image is split into $16 \times 16$ patches, resulting in 196 patches per image. 
        
        For images correctly classified by all models---100 images sampled from a pool of 200 to limit computational cost---saliency maps are generated for every evaluated XAI method. Starting from a fully masked baseline, patches are iteratively added in decreasing order of importance, while the model's predicted probability for the ground-truth class $y$ is recorded at each step, producing an \emph{insertion curve}. Conversely, starting from the original image, patches are progressively occluded in the same order to generate a \emph{deletion curve} (Fig.~\ref{insertioncurves}). 
        A high-quality saliency map should produce a rapid confidence rise under insertion---corresponding to a large area under the insertion curve---and a steep early confidence decrease under deletion---corresponding to a low area under the deletion curve. 
        To contextualize these results, each method is compared against a random baseline in which patches are inserted or deleted in uniformly random order, averaged across five permutations per image. 
        As reported in Appendix~\ref{baselinesinsertion}, all four XAI methods consistently surpass the baseline across all architectures, providing quantitative evidence that the generated attributions capture discriminative regions genuinely used during inference.    
        \begin{figure}[H]
            \centering
            \begin{subfigure}{\linewidth}
                \centering                \includegraphics[width=\linewidth]{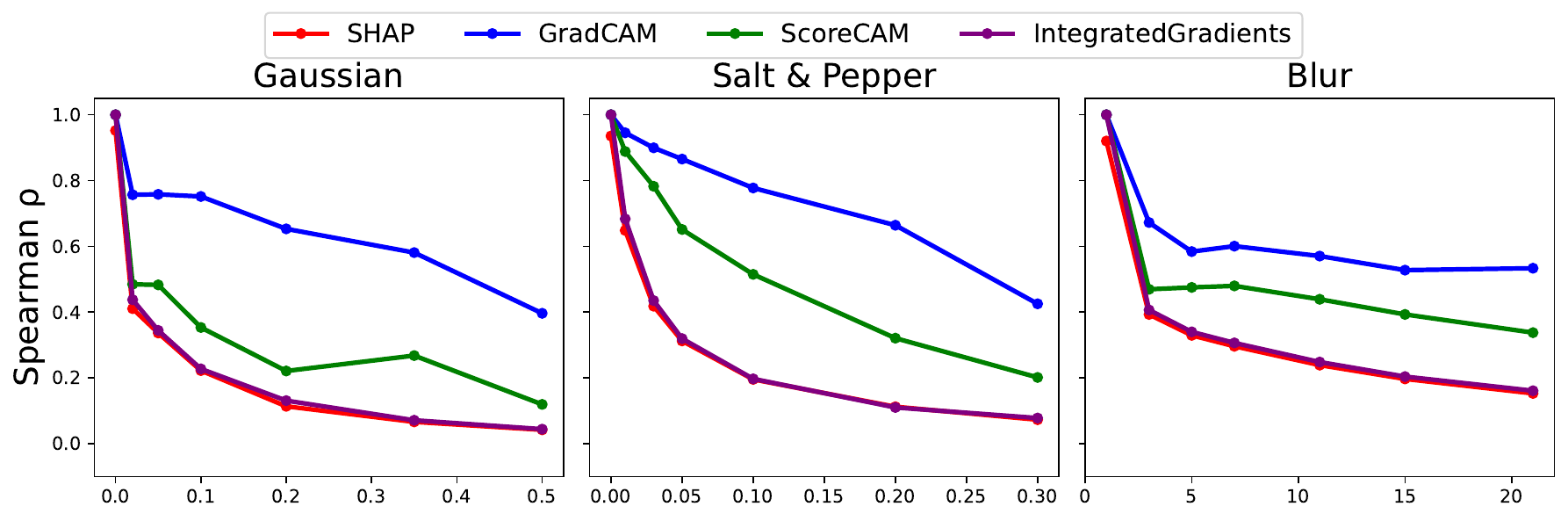}
                \caption{MobileNet model.}
            \end{subfigure}
        
            \begin{subfigure}{\linewidth}
                \centering                \includegraphics[width=\linewidth]{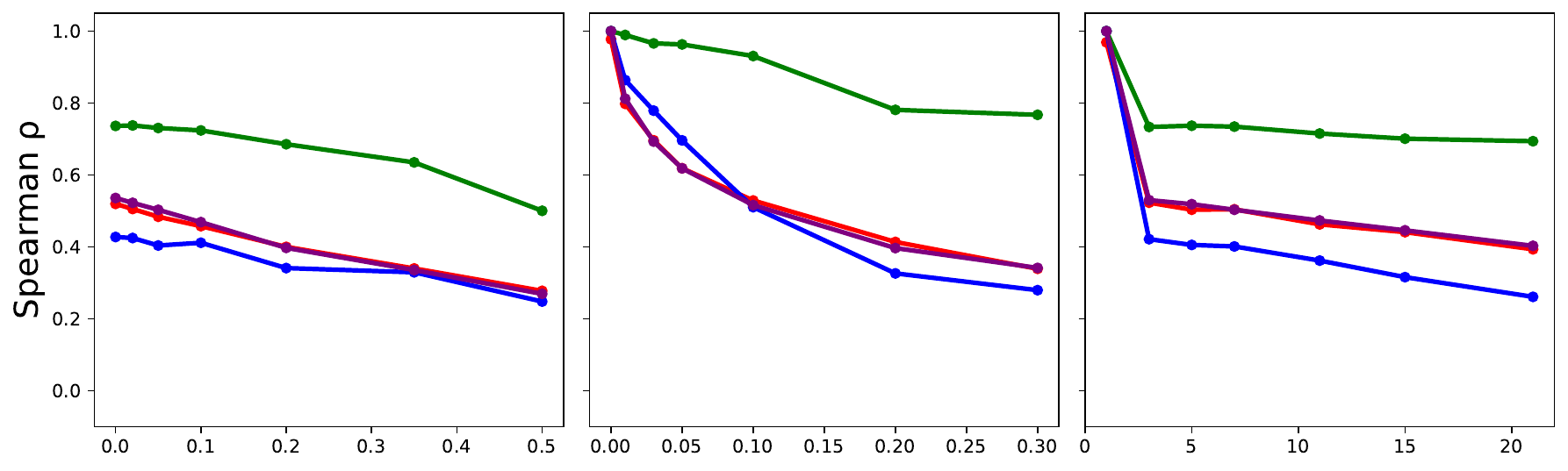}
                \caption{ResNet model.}
            \end{subfigure}
            
            \begin{subfigure}{\linewidth}
                \centering                \includegraphics[width=\linewidth]{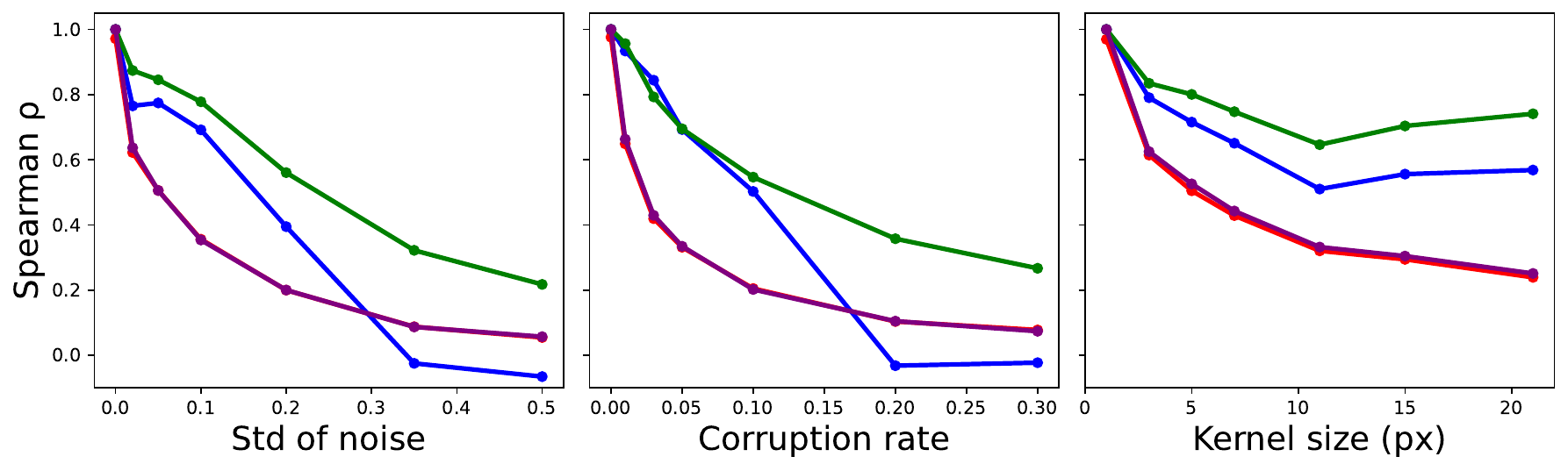}
                \caption{Efficient model.}
            \end{subfigure}
            \caption{Stability of explanations under perturbations.}
            \label{saliency_stability}
        \end{figure}
        Quantitatively, Figure~\ref{saliency_stability} reports $\rho$ between clean and corrupted Grad-CAM explanations across all architectures and corruption severities. Explanation stability closely follows classification confidence: $\rho$ remains above $90\%$ under mild perturbations but progressively decreases as corruption severity increases, suggesting that saliency-based explanations remain reliable in low-noise regimes---as shown in Figure\,\ref{corruptionxai}---while degrading alongside predictive performance. 
        Under mild Gaussian noise ($\sigma = 0.05$), Grad-CAM preserves the salient region found in the clean image, albeit with reduced activation intensity and broader spatial spread.
        Under severe corruption ($\sigma = 0.5$), attributions shift substantially away from the originally highlighted region---consistent with the qualitative result (Fig.\,\ref{corruptionxai})---, indicating a breakdown in explanation consistency. A similar trend is observed under salt-and-pepper noise.
        Additional models and corruption types are provided in Appendix~\ref{corruptionXAI}.

\section{Discussion}
    \label{discussion}
    This study examined four deep learning models spanning diverse design paradigms and model capacities, differing not only in predictive performance, but also in visual representation and computational requirements. The results provide several insights into the interplay between accuracy, efficiency, robustness, and interpretability in a clinically relevant deployment setting.
    
    \paragraph{Predictive performance.} All four architectures achieved strong classification performance, with MobileNetV3 achieving the highest accuracy. Despite minor numerical differences across metrics, the Friedman Omnibus test confirmed no statistically significant performance differences among the models at $(\alpha = 0.01)$. This finding suggests that architectures spanning substantially different parameter scales can converge to statistically comparable diagnostic performance. Consequently, lightweight architectures such as MobileNet and EfficientNet-B0 offer meaningful deployment advantage in resource-constrained settings without measurable loss in predictive performance.

    \paragraph{Accuracy vs.\ Efficiency.} The inclusion of efficient-by-design architectures alongside larger models was motivated by the need to examine the accuracy-efficiency frontier in low-resource clinical setting. The results are notable: MobileNetV3 achieved the highest cross-validated accuracy (96.35\%) and the highest AUC (0.992) among individual architectures, despite having substantially fewer parameters than ViT-B/16. EfficientNet-B0 achieved comparable performance while maintaining a similarly compact parameter footprint. These findings align with prior literature demonstrating that efficient-by-design architectures---through mechanisms such as depthwise separable convolutions and  compound scaling---can match or even substantially exceed larger architectures on domain-specific medical imaging tasks \cite{diagnostics15192515}, reinforcing their suitability for deployment under hardware-constraints.

    \paragraph{Ensembling classifiers.} The logit-level ensemble achieved the strongest overall performance across all reported metrics. Logit aggregation was preferred over majority voting because majority voting provides no principled tie-breaking mechanism for an even number of models, whereas averaging in logit space preserves the full confidence distribution of individual classifiers. Ensemble methods have consistently demonstrated improved performance across medical imaging tasks~\cite{supriyadi2025systematic}, and the present results further indicate that ensemble gains remain achievable even when the considered models are not individually statistically distinguishable.

    \paragraph{Noise influence on predictive performance.} Across all architectures and corruption types, prediction confidence decreases faster than classification accuracy, revealing a consistent decoupling between confidence and correctness. This behavior suggests that models become uncertain before substantial predictive degradation occurs, which is clinically valuable because uncertainty may provide an implicit signal that predictions are approaching the model’s reliability boundary.  In contrast, confidently incorrect predictions represent a substantially greater deployment risk. The observed confidence-accuracy gap therefore reflects more than a statistical artifact and may instead served as a practical proxy for model reliability under distribution shift.

    \paragraph{Architecture-specific robustness patterns.} ResNet-18 demonstrates the strongest resistance to salt-and-pepper noise, maintaining perfect accuracy through the first three corruption levels. This robustness likely arises from residual connections, that preserve low-frequency structural information despite high-frequency pixel-level corruption \cite{he2016deep}, making the architecture particularly suitable for deployment with low-cost microscopy hardware. 
    EfficientNet-B0 exhibits the greatest sensitivity to Gaussian noise, with accuracy approaching approximately 50\% at the highest severity. This vulnerability may stem from the compound scaling strategy, which optimizes fine-grained spatial statistics that directly disrupted by Gaussian perturbations \cite{tan2019efficientnet}. 
    MobileNetV3 exhibits moderate and comparatively consistent degradation across all corruption types, suggesting that  depthwise separable convolution provide implicit regularization without explicitly learning  strongly noise-invariant representations. 
    ViT-B/16 displays a non-monotonic accuracy response under Gaussian noise, partially recovering at intermediate severities before collapsing at highest levels. This behavior may reflect the self-attention mechanism's ability to average local corruptions perturbations under moderate noise before heavily corrupted patches begin dominating the learned attention distribution. 
    Blur is comparatively well tolerated across all architectures, as it preserves the coarse spatial structure of the parasite while removing high-frequency texture information that models appear to treat as less discriminative. Consequently, optical defocus represents a less severe degradation source than stochastic sensor noise or transmission artifacts.    
    
    \paragraph{Explainability.} A major contribution of this study is the joint evaluation of predictive performance and post-hoc explanation quality. ViT-B/16 was excluded from the explainability analysis because the selected XAI methods were originally designed for convolutional architectures. Grad-CAM and Score-CAM produce coarse, class-discriminative localization maps that consistently emphasize broad cellular regions, particularly near the center of parasitized erythrocytes---a pattern consistent with the spatial distribution of pigment granules and nuclear material within infected cells \cite{pan2009survey}. 
    Integrated Gradients and SHAP instead generate finer-grained and spatially precise attributions. 
    For ResNet-18, both fine-grained methods highlight broader and less targeted regions compared with EfficientNet-B0 and MobileNetV3, suggesting that the latter architectures learn more spatially compact and diagnostically relevant representations. These findings further support the explainability potential of lightweight architectures, as both Integrated-Gradients and SHAP consistently assign the highest attribution weights to parasite-localized regions, particularly emphasis around the characteristic ring-stage morphology associated with malaria diagnosis \cite{pan2009survey}.
    
    \paragraph{Architecture-specific explanation patterns.} Pairwise disagreement analysis reveals architecturally distinct spatial patterns. For EfficientNet-B0, disagreement between Grad-CAM and Score-CAM is concentrated near the parasite membrane boundary, while ResNet-18 disagreement is localized within the parasite interior, suggesting stronger reliance on internal chromatin structure. MobileNetV3 exhibits the most fine-grained disagreement structure, spanning both parasite boundary and internal sub-regions, reflecting its tendency toward highly localized spatial attribution. Importantly, these disagreement patterns emerge independently of prediction correctness, indicating that network architecture implicitly shapes where explanation uncertainty appears. This finding has direct implications for the trustworthiness and interpretability of post-hoc XAI methods in clinical settings.
    
    \paragraph{Explanation robustness to perturbation.} One of the key findings of this study is that none of the evaluated XAI methods remains robust under substantial input corruption. Explanation stability degrades consistently with increasing perturbation severity across all architectures and explanation methods, in some cases exponentially. With the exception of ResNet-18, all architectures exhibit declining Spearman correlations even under moderate Gaussian noise. 
    Crucially, this degradation occurs at corruption levels that remain plausible in low-resource clinical environments, while predictive accuracy often remains comparatively stable. This decoupling is concerning as clinicians may continue trusting visually coherent saliency maps even when the underlying explanation mechanism has become unreliable.
    The contrasting behavior under blur provides  additional insight. Unlike stochastic noise, which disrupts the high-frequency features, blur acts primarily as a low-pass filter that preserves coarse spatial structure while progressively reducing activation magnitude. The relative stability of explanations under blur therefore suggests that localization relies predominantly on low-to-mid frequency spatial features, consistent with the coarse morphological cues relevant to parasite detection \cite{fitri2022malaria}. This property may incidentally confer robustness to optical defocus artifact commonly encountered in resource-resource microscopy workflows.

    \paragraph{Ethical implications of explanation unreliability.} The observation that post-hoc saliency maps degrade under clinically plausible corruption levels---even when predictive accuracy remains high---introduces important ethical concerns for deployment. A clinician may receive a confident and accurate prediction, along with a visually plausible but corrupted explanation map, without any reliable indication that the explanation itself is misleading. This creates an asymmetric failure mode in which the model appears trustworthy precisely when the explanation is silently wrong. 
    Combined with the previously observed confidence-accuracy decoupling, these findings suggest that neither prediction confidence nor post-hoc explanations alone constitute sufficient safeguards for clinical oversight. Deployment pipelines in resource-constrained environments should therefore treat low prediction confidence and high input corruption---potentially detectable through lightweight noise-estimation procedures---as complementary triggers for mandatory human review, rather than relying on saliency maps as a secondary validation mechanism under adverse imaging conditions.

    \paragraph{Generalisability limits and equity implications.} Several limitations constrain the generalizability of the present findings. Experiments were conducted exclusively on the NLM-Malaria dataset, which contains a single Plasmodium species (emph{P. falciparum) acquired under standardized staining and magnification conditions at a single institution. However, malaria-endemic regions in Sub-Saharan Africa frequently encounter substantially different diagnostic conditions, including variable staining quality, multiple co-circulating \textit{Plasmodium} species, lower-cost microscopes, and operator-dependent slide preparation procedures \cite{wongsrichanalai2007review}. Consequently, the populations most likely to benefit from automated diagnosis remain underrepresented in the current validation setting. 
    Practical deployment would additionally require an upstream cell-detection stage, since real clinical workflows operate on whole-slide microscopy images rather than pre-segmented individual cells. Furthermore, the dataset contains no demographic and acquisition metadata, preventing analysis of performance across patient groups, parasite developmental stages, or imaging conditions---factors essential for equitable deployment assessment. 
    Addressing these limitations requires external validation using African clinical datasets and field-collected microscopy data. Potential candidate datasets include the  \cite{yu2020malaria}, which incorporates multiple \textit{Plasmodium} species, as well as prospective data collection in endemic-regions or the Lacuna Malaria dataset covering thin and thick smear cell images from Ghana and Uganda \cite{DVN_VEADSE_2023}.  
    Until such validation is performed, the deployment recommendations presented here should be interpreted as technically motivated hypotheses rather than clinically validated guidelines.

\section{Conclusion}
    This work investigated four deep learning architectures for automated malaria diagnosis through a joint evaluation of predictive performance, robustness, computational efficiency, and post-hoc explainability in low-resource clinical settings.
    Across all reported metrics, lightweight architectures such as MobileNetV3 and EfficientNet-B0 achieved performance comparable to substantially larger models, supporting the practical relevance of efficient-by-design architectures for resource-constrained deployment. 
    The study additionally demonstrated that robustness and explainability should be evaluated jointly rather than independently. Although predictive accuracy remained relatively stable under moderate corruption, explanation reliability degraded substantially across all evaluated XAI methods, revealing a critical limitation of post-hoc saliency-based validation under distribution shift. The observed decoupling between prediction confidence and accuracy further suggests that uncertainty signals may provide useful safeguards for human oversight in noisy clinical environments.

    Overall, the findings highlight that  that model selection for medical imaging should not rely solely on predictive performance, but should additionally consider computational efficiency, robustness, and explanation stability under realistic acquisition conditions. 
    Future work may extend the present analysis to transformer-based explainability methods as well as inherently interpretable architectures, including sparse BagNet \cite{donteu2023sparse} and explainable hybrid CNN-Transformer \cite{djoumessi2025hybrid}, whose predictions are directly grounded in localized image evidence rather than relying on post-hoc attribution mechanisms.
    

\section*{Ethical Statement}  
    This study uses the publicly available NLM-Malaria dataset, which was released without patient-identifiable information No new data were collected, and no human subjects were directly involved in this research. The ethical implications of explanation unreliability under image corruption, together with the generalizability limitations of the NLM-Malaria dataset and their equity implications for African clinical deployment, are discussed in Section \ref{discussion}. Additional ethical considerations are outlined below.

    \paragraph{Misdiagnosis risk.} Automated malaria diagnosis systems inherently carry the risk of false negatives---classifying an infected cell as healthy--which may delay treatment and contribute to severe disease progression or mortality, particularly among children under five in resource-limited clinical settings. Consequently, the models evaluated in this work should be considered decision-support tools for trained clinicians rather than replacements for expert review. Clinical deployment should additionally incorporate clearly defined confidence thresholds below which mandatory human verification is required.

    \paragraph{Responsible deployment.} Lightweight architectures are identified as promising candidates for deployment in resource-constrained African clinical environments. However, technical validation on a controlled benchmark dataset alone is insufficient for real-world clinical adoption. Prospective validation in target clinical environments, collaboration with local clinicians and health authorities, and compliance with national medical device regulations remain necessary prerequisites that fall beyond the scope of the present study.

\section*{Acknowledgments}
    This project was supported by the Hertie Foundation and AIMS Rwanda. 
    The authors thank Yvan Guifo for hosting the project.

\bibliographystyle{named}
\bibliography{ijcai26}

\newpage

\appendix
\section{Individual Implementation Results}
\label{appendix}
\subsection{ResNet-18}

\begin{table}[H]
\centering
\caption{Performance metrics.}
\begin{tabular}{lcccc}
\toprule
\textbf{Metric} & \textbf{Mean} & \textbf{Std} & \textbf{Lower CI} & \textbf{Upper CI} \\
\midrule
Accuracy  & 0.9614 & 0.0037 & 0.9568& 0.9660 \\
Precision & 0.9618 & 0.0038 & 0.9571& 0.9665 \\
Recall    & 0.9614 & 0.0037 & 0.9568& 0.9660 \\
F1-score  & 0.9614 & 0.0037 & 0.9568& 0.9660 \\
AUC       & 0.9907 & 0.0020 & 0.9881& 0.9932 \\
\bottomrule
\end{tabular}
\label{tab:cv_metrics}
\end{table}

\begin{figure}[H]
    \centering
    \begin{subfigure}[t]{0.48\columnwidth}
        \includegraphics[width=\linewidth]{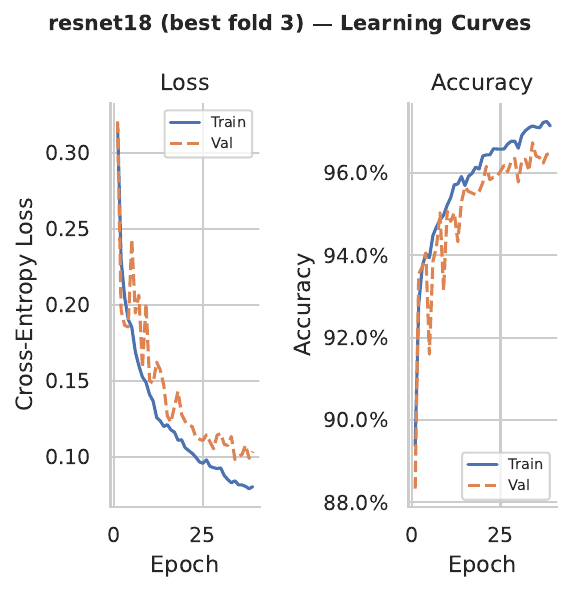}
        \caption{Best learning curves}
    \end{subfigure}
    \hfill
    \begin{subfigure}[t]{0.48\columnwidth}
        \includegraphics[width=\linewidth]{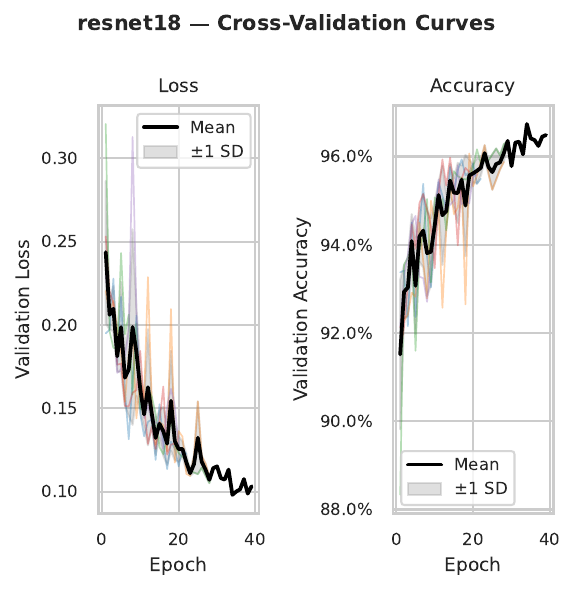}
        \caption{Cross-validation curves}
    \end{subfigure}


    \caption{ResNet-18 learning curves.}
    \label{fig:resnet_2}
\end{figure}


\subsection{EfficientNet\_b0}

\begin{table}[H]
\centering
\caption{Performance Metrics}
\begin{tabular}{lccccc}
\toprule
\textbf{Metric} & \textbf{Mean} & \textbf{Std} & \textbf{Lower CI} & \textbf{Upper CI} \\
\midrule
Accuracy  & 0.9623 & 0.0035 & 0.9580 & 0.9667 \\
Precision & 0.9626 & 0.0035 & 0.9583 & 0.9669 \\
Recall    & 0.9623 & 0.0035 & 0.9580 & 0.9667 \\
F1 Score  & 0.9623 & 0.0035 & 0.9580 & 0.9667 \\
AUC       & 0.9923 & 0.0019 & 0.9899 & 0.9947 \\
\bottomrule
\end{tabular}
\label{tab:model_metrics_3}
\end{table}

\begin{figure}[H]
    \centering
    \begin{subfigure}[t]{0.48\columnwidth}
        \includegraphics[width=\linewidth]{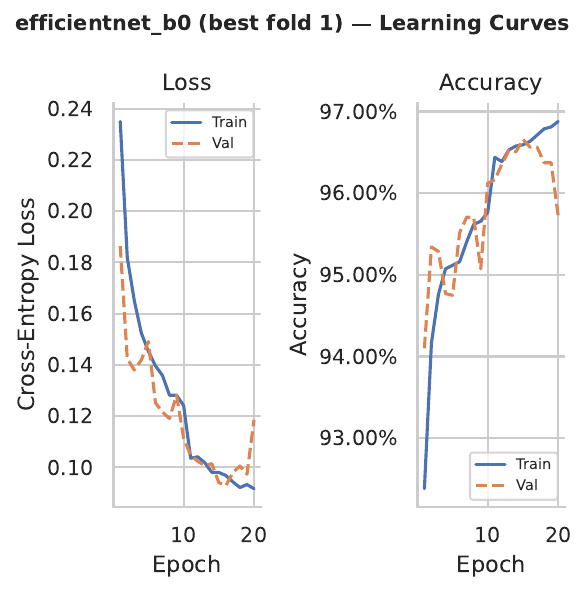}
        \caption{Best learning curves}
    \end{subfigure}
    \hfill
    \begin{subfigure}[t]{0.48\columnwidth}
        \includegraphics[width=\linewidth]{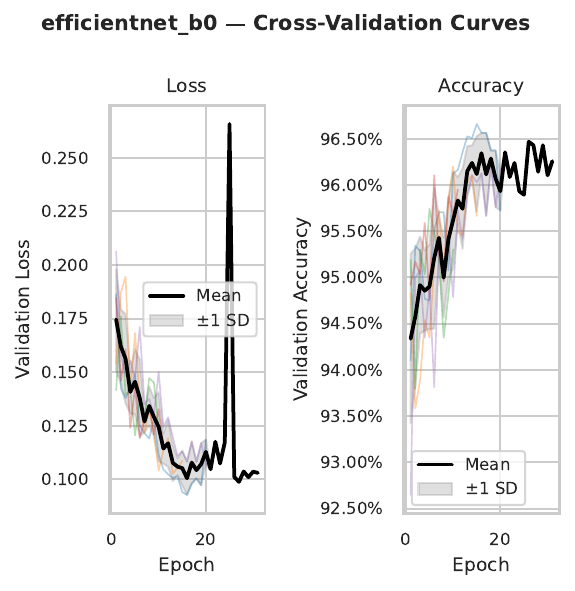}
        \caption{Cross-validation curves}
    \end{subfigure}

    \caption{EfficientNet learning curves.}
    \label{fig:resnet}
\end{figure}


\subsection{MobileNet\_v3\_large}

\begin{table}[H]
\centering
\caption{Performance Metrics}
\begin{tabular}{lccccc}
\toprule
\textbf{Metric} & \textbf{Mean} & \textbf{Std} & \textbf{Lower CI} & \textbf{Upper CI} \\
\midrule
Accuracy  & 0.9635 & 0.0055 & 0.9567 & 0.9703 \\
Precision & 0.9637 & 0.0054 & 0.9570 & 0.9703 \\
Recall    & 0.9635 & 0.0055 & 0.9567 & 0.9703 \\
F1 Score  & 0.9635 & 0.0055 & 0.9567 & 0.9703 \\
AUC       & 0.9922 & 0.0026 & 0.9890 & 0.9954 \\
\bottomrule
\end{tabular}
\label{tab:model_metrics}
\end{table}

\begin{figure}[H]
    \centering
    \begin{subfigure}[t]{0.48\columnwidth}
        \includegraphics[width=\linewidth]{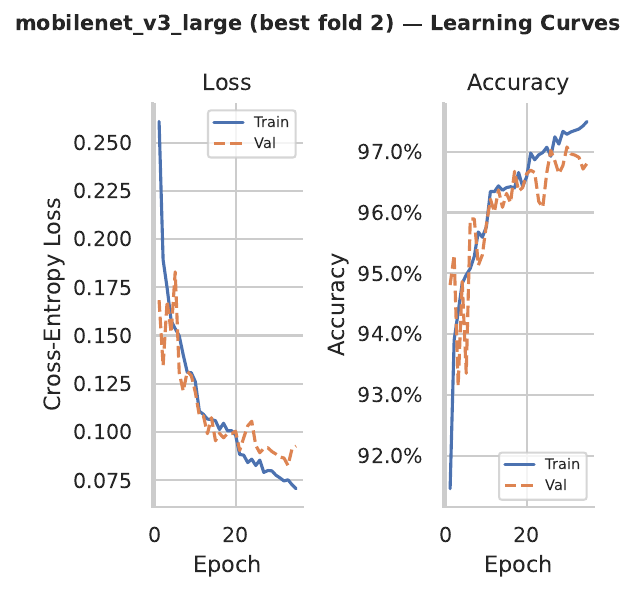}
        \caption{Best learning curves}
    \end{subfigure}
    \hfill
    \begin{subfigure}[t]{0.48\columnwidth}
        \includegraphics[width=\linewidth]{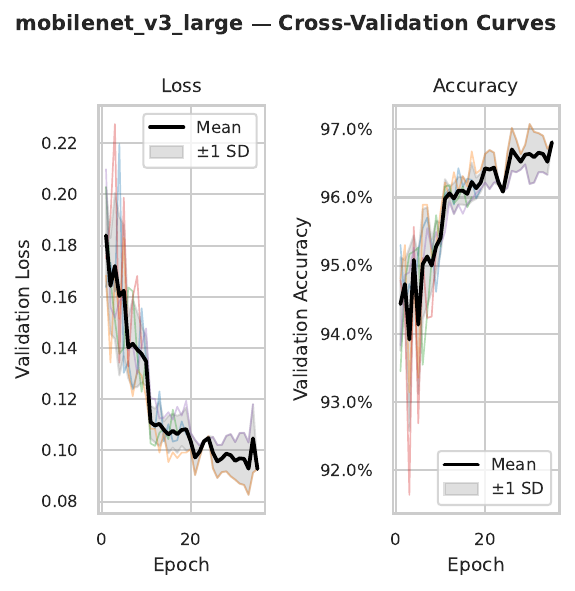}
        \caption{Cross-validation curves}
    \end{subfigure}


    \caption{MobileNet learning curves}
    \label{vit_}
\end{figure}


\subsection{ViT\_b\_16}

\begin{table}[H]
\centering
\caption{Performance Metrics}
\begin{tabular}{lccccc}
\toprule
\textbf{Metric} & \textbf{Mean} & \textbf{Std} & \textbf{Lower CI} & \textbf{Upper CI} \\
\midrule
Accuracy  & 0.9581 & 0.0045 & 0.9525 & 0.9638 \\
Precision & 0.9583 & 0.0045 & 0.9528 & 0.9639 \\
Recall    & 0.9581 & 0.0045 & 0.9525 & 0.9638 \\
F1 Score  & 0.9581 & 0.0045 & 0.9525 & 0.9638 \\
AUC       & 0.9889 & 0.0026 & 0.9857 & 0.9921 \\
\bottomrule
\end{tabular}
\label{tab:model_metrics_2}
\end{table}

\begin{figure}[H]
    \centering
    \begin{subfigure}[t]{0.48\columnwidth}
        \includegraphics[width=\linewidth]{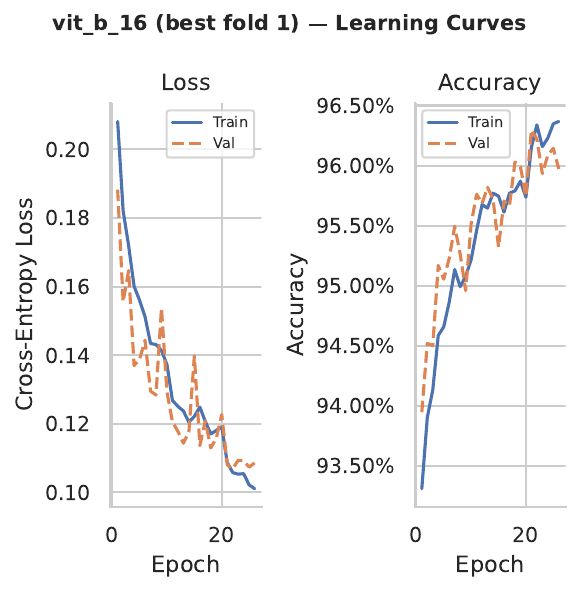}
        \caption{Best learning curves }
    \end{subfigure}
    \hfill
    \begin{subfigure}[t]{0.48\columnwidth}
        \includegraphics[width=\linewidth]{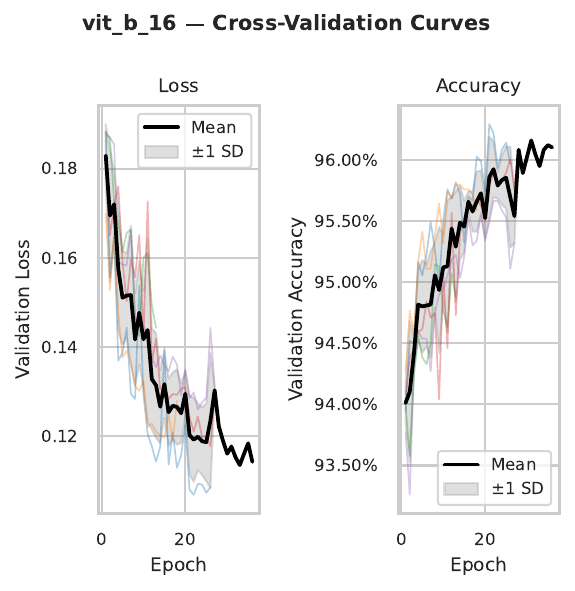}
        \caption{Cross-validation curves}
    \end{subfigure}

    \caption{ViT learning curves}
    \label{vit}
\end{figure}

\section{Robustness Analysis on Corrupted Images}
    \label{sensitivityAppendix}
    The following tables highlight the detailed results of the model's sensitivity under different noise types. They show how accuracy and confidence reduce as we increase severity level. Looking at the worst-case scenarios: under Gaussian noise with $\sigma = 0.5$, EfficientNet performs the poorest, with accuracy dropping to 51\%. Under blur, where all models are not highly affected, EfficientNet similarly drops to 83\% in accuracy --- meaning it misclassified 17\% of the cells originally predicted correctly before noise was applied. The same pattern holds for salt-and-pepper noise, where EfficientNet again performs the poorest, dropping to 63\%.

\begin{table}[h]
\centering
\caption{Results under Gaussian noise}
\setlength{\tabcolsep}{3pt}
\begin{tabular}{l|cc|cc|cc|cc}
\hline
Level &
\multicolumn{2}{c|}{ResNet} &
\multicolumn{2}{c|}{MobileNet} &
\multicolumn{2}{c|}{EfficientNet} &
\multicolumn{2}{c}{ViT} \\
& Acc. & Conf. & Acc. & Conf. & Acc. & Conf. & Acc. & Conf. \\
\hline
0.00 & 1.000 & 0.975 & 1.000 & 0.974 & 1.000 & 0.965 & 1.000 & 0.971 \\ 
0.02 & 0.890 & 0.901 & 0.960 & 0.923 & 0.960 & 0.926 & 0.880 & 0.904 \\ 
0.05 & 0.910 & 0.900 & 0.960 & 0.929 & 0.940 & 0.922 & 0.870 & 0.903 \\ 
0.10 & 0.900 & 0.902 & 0.950 & 0.915 & 0.980 & 0.910 & 0.920 & 0.898 \\ 
0.20 & 0.880 & 0.875 & 0.880 & 0.854 & 0.880 & 0.809 & 0.930 & 0.863 \\ 
0.35 & 0.820 & 0.860 & 0.750 & 0.786 & 0.590 & 0.678 & 0.780 & 0.779 \\ 
0.50 & 0.730 & 0.854 & 0.700 & 0.737 & 0.510 & 0.687 & 0.570 & 0.826 \\ 

\hline
\end{tabular}
\end{table}

\begin{table}[h]
\centering
\caption{Results under Blur noise}
\setlength{\tabcolsep}{3pt}
\begin{tabular}{l|cc|cc|cc|cc}
\hline
Level &
\multicolumn{2}{c|}{ResNet} &
\multicolumn{2}{c|}{MobileNet} &
\multicolumn{2}{c|}{EfficientNet} &
\multicolumn{2}{c}{ViT} \\
& Acc. & Conf. & Acc. & Conf. & Acc. & Conf. & Acc. & Conf. \\
\hline
1.00 & 1.000 & 0.975 & 1.000 & 0.974 & 1.000 & 0.965 & 1.000 & 0.971 \\ 
3.00 & 0.920 & 0.902 & 0.930 & 0.862 & 0.940 & 0.894 & 0.890 & 0.908 \\ 
5.00 & 0.910 & 0.903 & 0.940 & 0.833 & 0.920 & 0.848 & 0.900 & 0.910 \\ 
7.00 & 0.910 & 0.902 & 0.930 & 0.840 & 0.860 & 0.825 & 0.900 & 0.908 \\ 
11.00 & 0.910 & 0.900 & 0.930 & 0.833 & 0.780 & 0.770 & 0.930 & 0.902 \\ 
15.00 & 0.910 & 0.902 & 0.910 & 0.818 & 0.800 & 0.800 & 0.910 & 0.902 \\ 
21.00 & 0.910 & 0.897 & 0.890 & 0.804 & 0.830 & 0.838 & 0.870 & 0.899 \\ 

\hline
\end{tabular}
\end{table}

\begin{table}[h]
\centering
\caption{Results under Salt \& Pepper noise}
\setlength{\tabcolsep}{3pt}
\begin{tabular}{l|cc|cc|cc|cc}
\hline
Level &
\multicolumn{2}{c|}{ResNet} &
\multicolumn{2}{c|}{MobileNet} &
\multicolumn{2}{c|}{EfficientNet} &
\multicolumn{2}{c}{ViT} \\
& Acc. & Conf. & Acc. & Conf. & Acc. & Conf. & Acc. & Conf. \\
\hline
0.00 & 1.000 & 0.975 & 1.000 & 0.974 & 1.000 & 0.965 & 1.000 & 0.971 \\ 
0.01 & 1.000 & 0.969 & 1.000 & 0.970 & 1.000 & 0.942 & 1.000 & 0.970 \\ 
0.03 & 1.000 & 0.953 & 1.000 & 0.957 & 0.980 & 0.919 & 0.990 & 0.966 \\ 
0.05 & 1.000 & 0.934 & 0.980 & 0.948 & 1.000 & 0.870 & 0.990 & 0.956 \\ 
0.10 & 0.950 & 0.881 & 0.990 & 0.924 & 0.950 & 0.793 & 0.990 & 0.921 \\ 
0.20 & 0.910 & 0.840 & 0.990 & 0.882 & 0.740 & 0.688 & 0.890 & 0.847 \\ 
0.30 & 0.920 & 0.848 & 0.940 & 0.842 & 0.630 & 0.664 & 0.730 & 0.810 \\ 

\hline
\end{tabular}
\end{table}

\section{XAI disagreement and agreement maps}
    \label{disagreement}
    In the following agreement/disagreement maps, we vary the value of $k$. For the disagreement map, $k$ controls the threshold above which a pixel is flagged as a meaningful disagreement: a smaller $k$ (e.g., $k=0.5$) sets a lower threshold relative to the mean, capturing more pixels as disagreeing and revealing finer, more distributed regions of divergence; while a larger $k$ (e.g., $k=2$) raises the threshold, restricting flagged regions to only those with strong, statistically significant disagreement. For the agreement map, $k$ similarly controls how strictly each individual map must exceed its own mean before a pixel qualifies as salient in both --- a smaller $k$ is more permissive, while a larger $k$ demands stronger unilateral saliency from each method before their conjunction is taken.

\begin{figure}[H]
    \centering
    
    \begin{subfigure}{\linewidth}
        \centering
        \includegraphics[width=\linewidth]{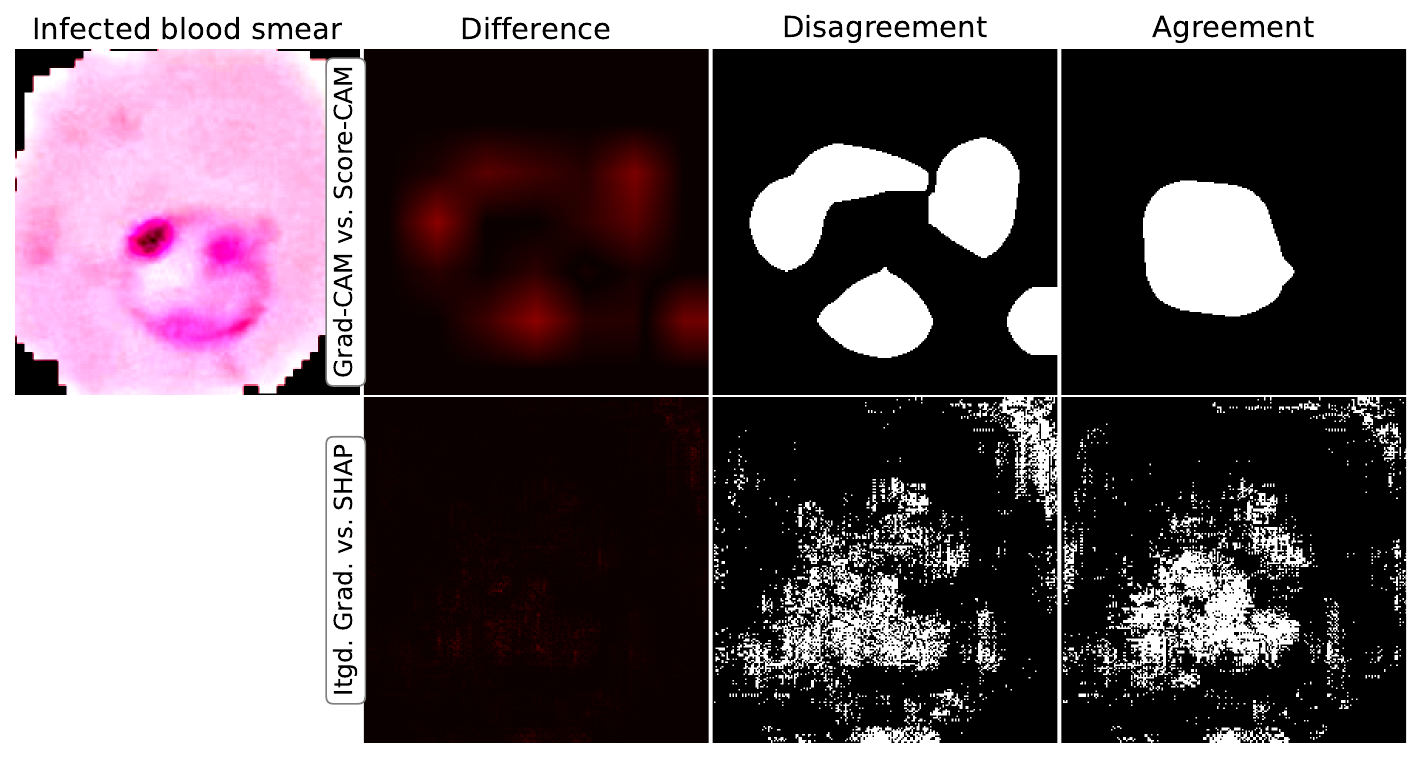}
        \caption{EfficientNet}
        \label{}
    \end{subfigure}
        
    \begin{subfigure}{\linewidth}
        \centering
        \includegraphics[width=\linewidth]{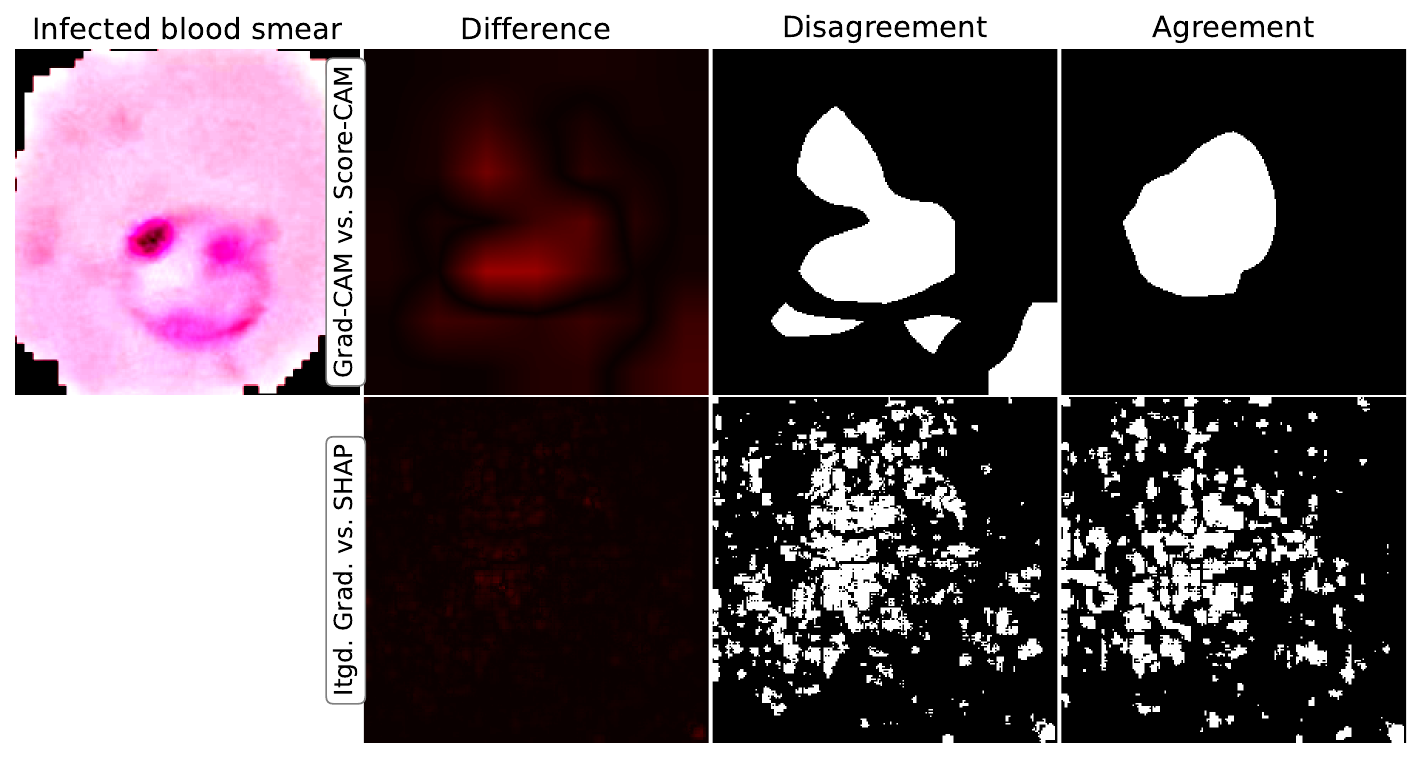}
        \caption{ResNet}
        \label{}
    \end{subfigure}
        
    \begin{subfigure}{\linewidth}
        \centering
        \includegraphics[width=\linewidth]{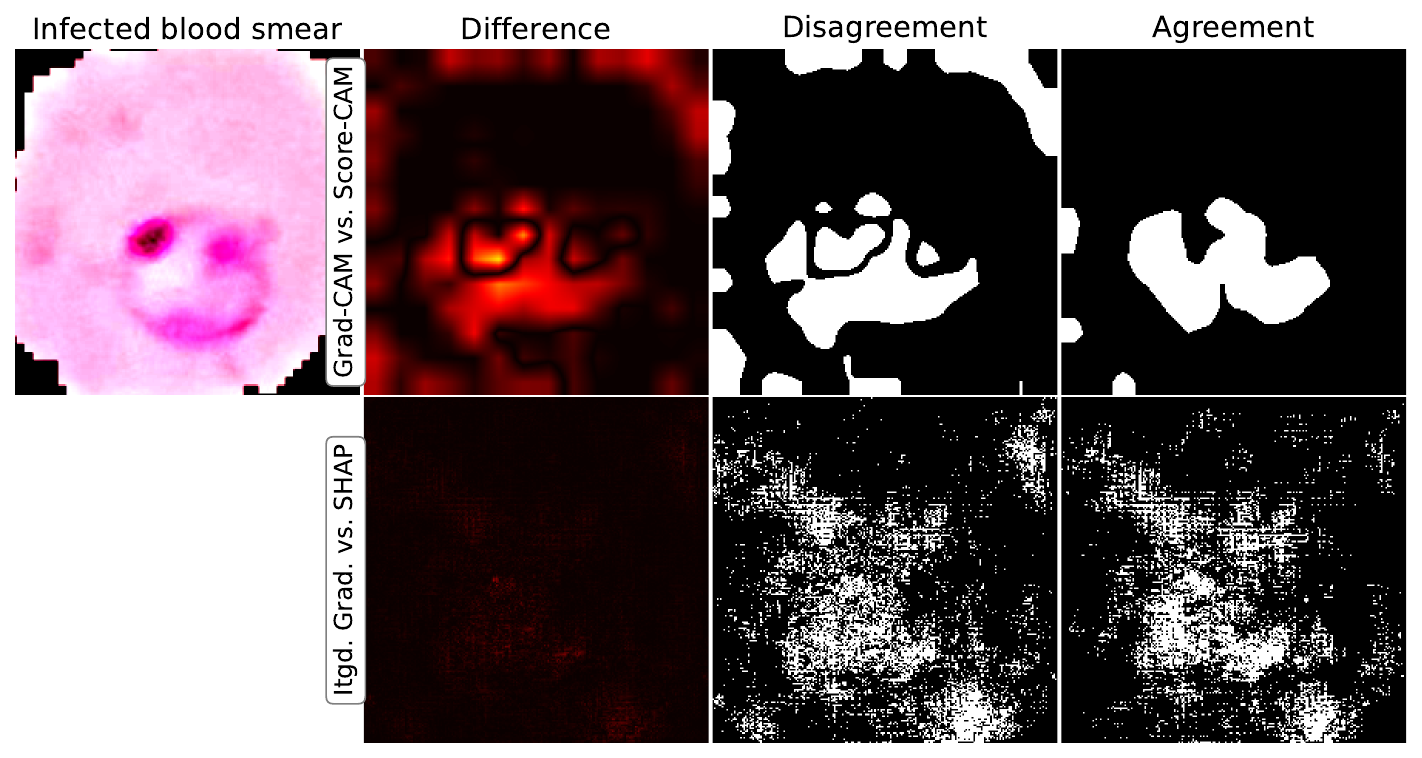}
        \caption{MobileNet}
        \label{}
    \end{subfigure}
    
    \caption{Disagreement maps across different model for $k = 0.5$}
    \label{disagreement_all_app}
\end{figure}

\begin{figure}[H]
    \centering
    
    \begin{subfigure}{\linewidth}
        \centering
        \includegraphics[width=\linewidth]{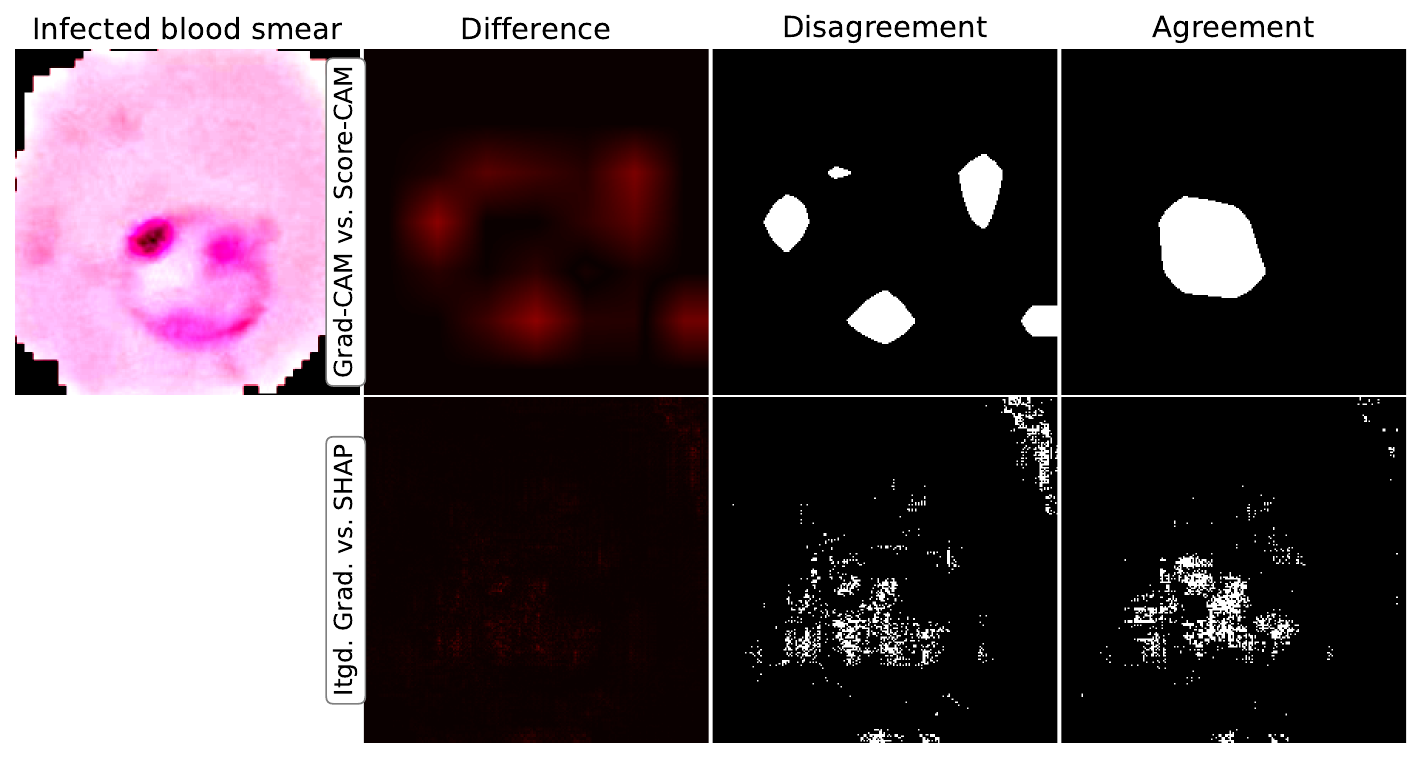}
        \caption{EfficientNet}
        \label{}
    \end{subfigure}
        
    \begin{subfigure}{\linewidth}
        \centering
        \includegraphics[width=\linewidth]{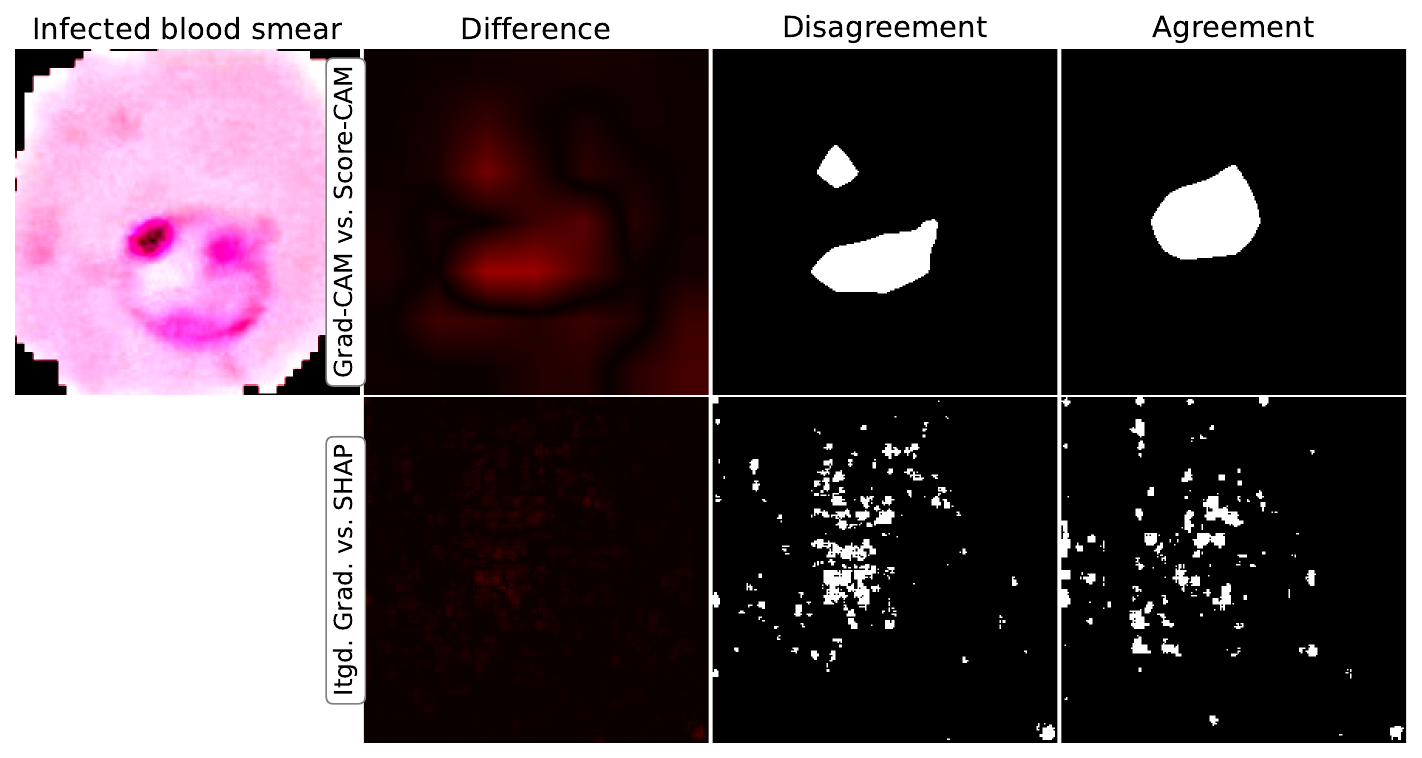}
        \caption{ResNet}
        \label{}
    \end{subfigure}
        
    \begin{subfigure}{\linewidth}
        \centering
        \includegraphics[width=\linewidth]{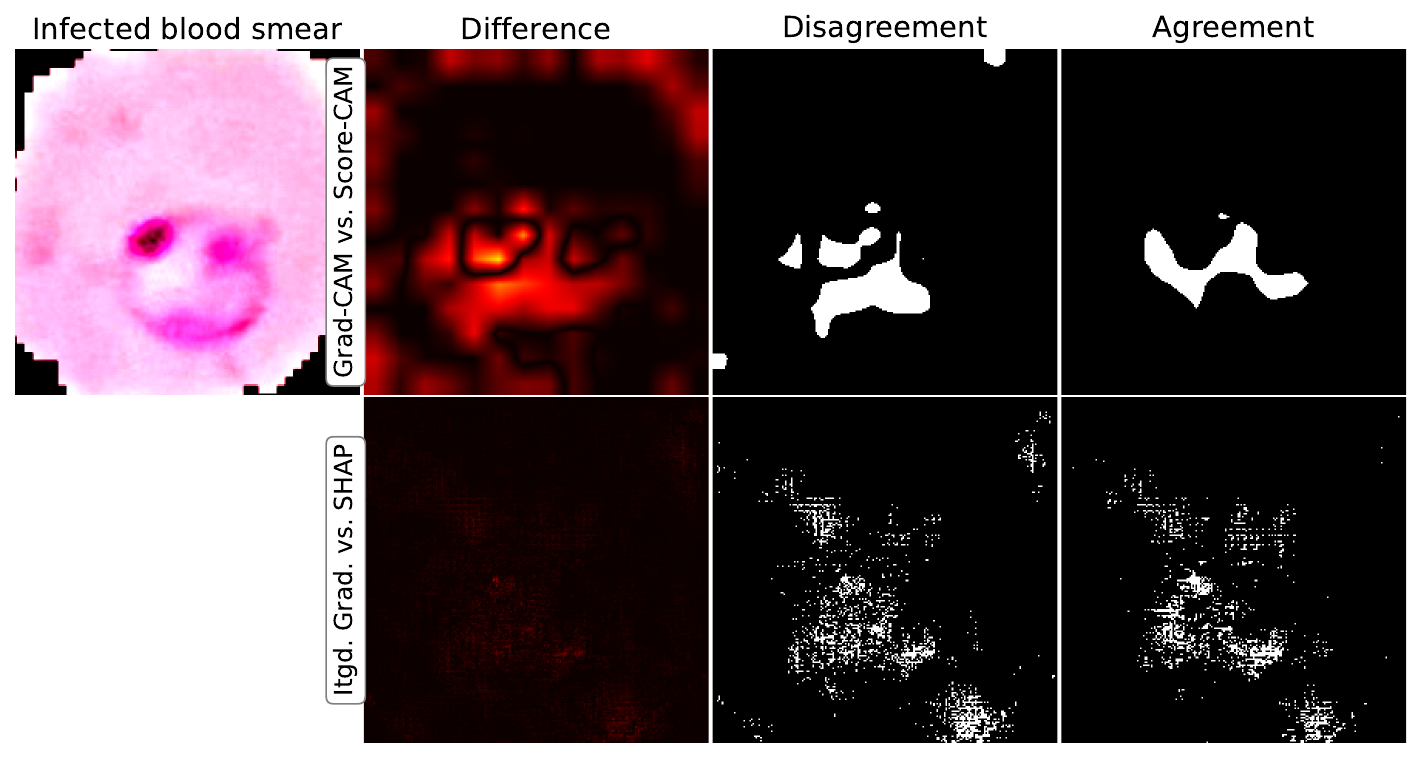}
        \caption{MobileNet}
        \label{}
    \end{subfigure}
    
    \caption{Disagreement maps across different model for $k = 2$}
    \label{}
\end{figure}


\section{Sensitivity of XAI Post-hoc methods under corruptions}
    \label{corruptionXAI}
    \begin{figure}[H]
        \centering
        \begin{subfigure}{\columnwidth}
            \centering
            \includegraphics[width=\linewidth]{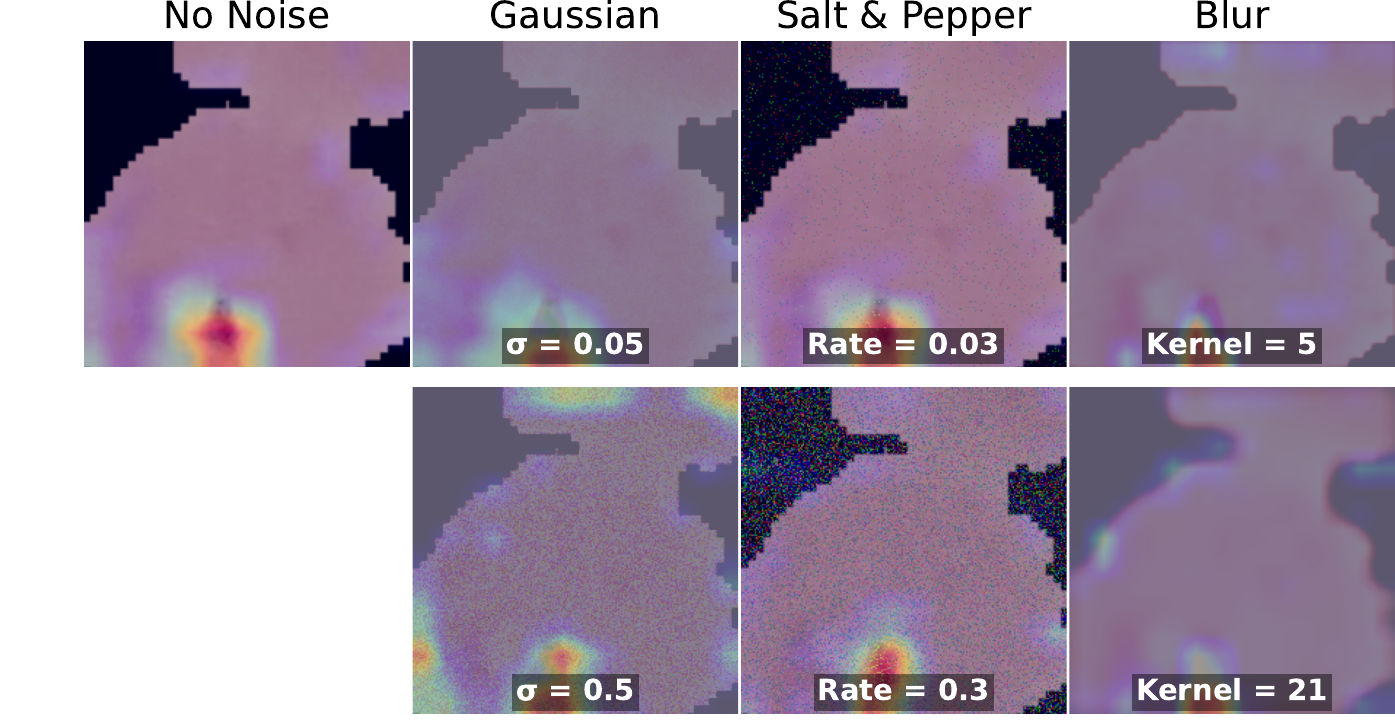}
            \caption{Grad-CAM}
        \end{subfigure}
        \hfill
        \begin{subfigure}{\columnwidth}
            \centering
            \includegraphics[width=\linewidth]{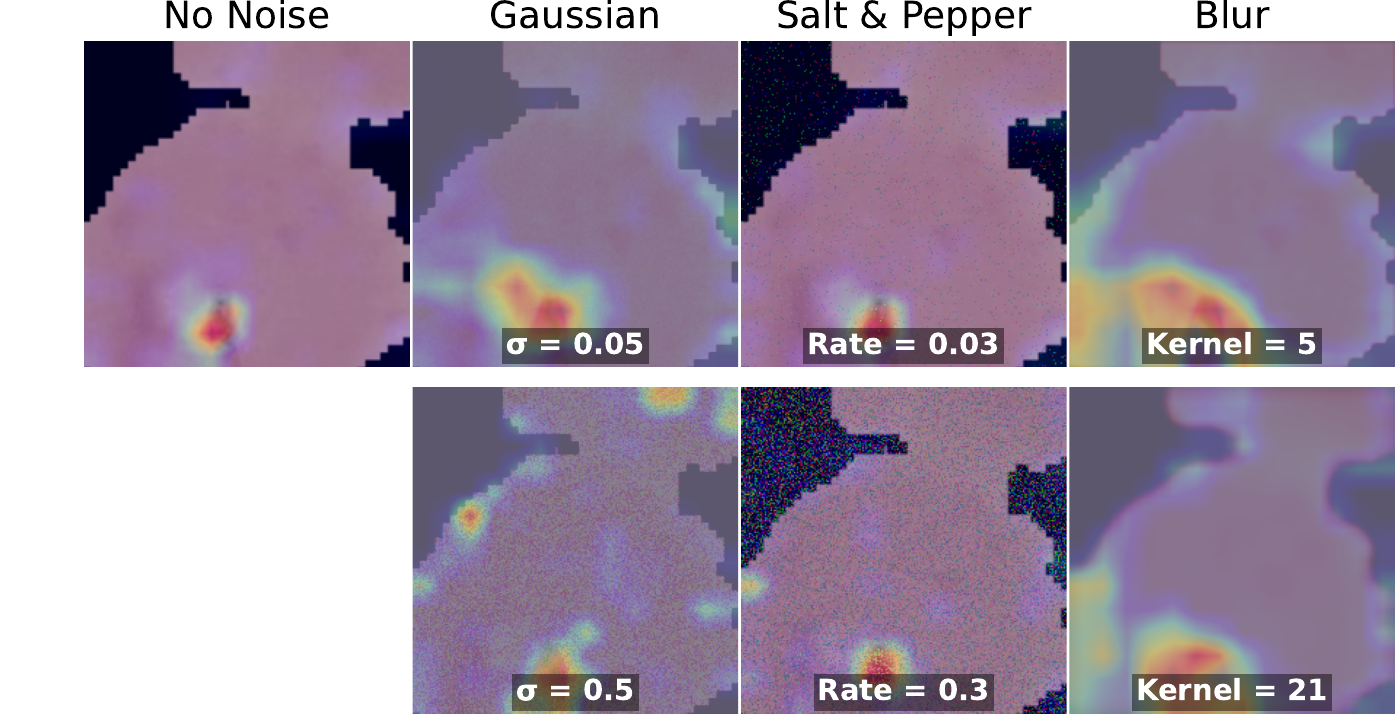}
            \caption{Score-CAM}
        \end{subfigure}
    
        \begin{subfigure}{\columnwidth}
            \centering
            \includegraphics[width=\linewidth]{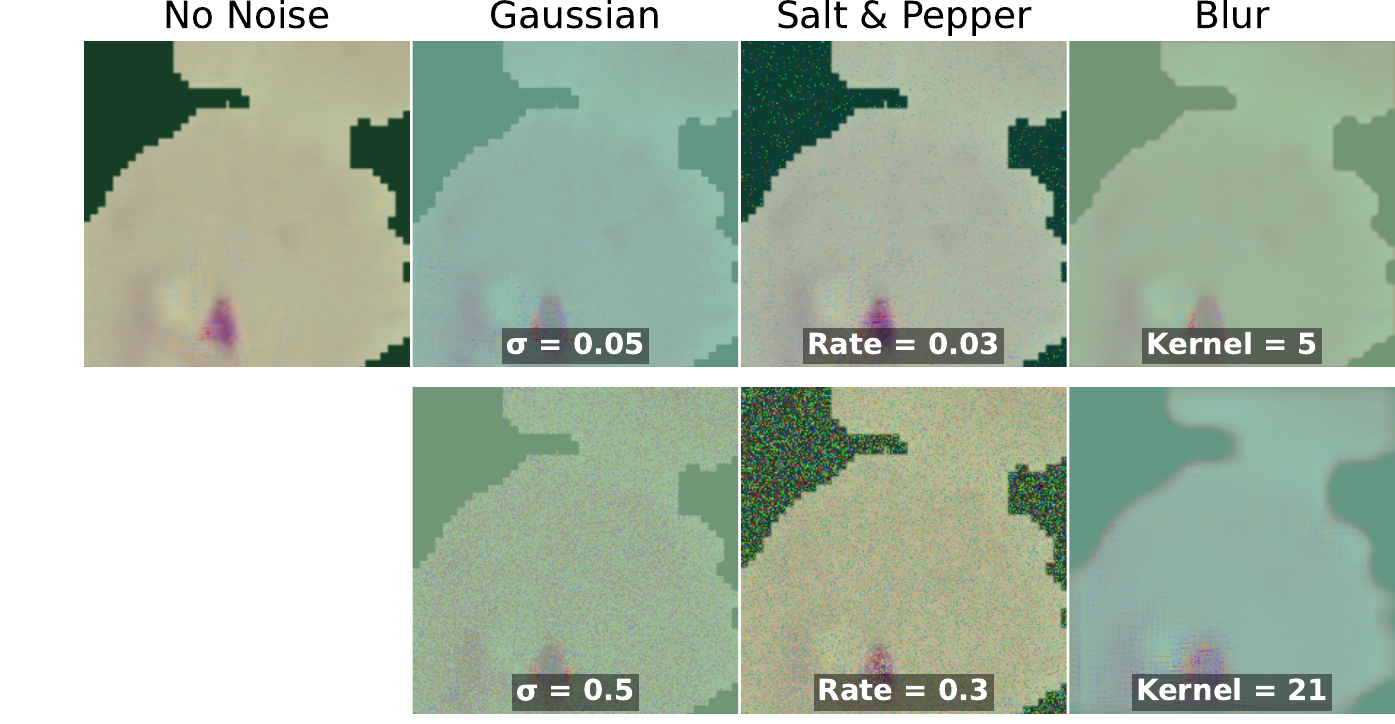}
            \caption{SHAP}
        \end{subfigure}
        \hfill
        \begin{subfigure}{\columnwidth}
            \centering
            \includegraphics[width=\linewidth]{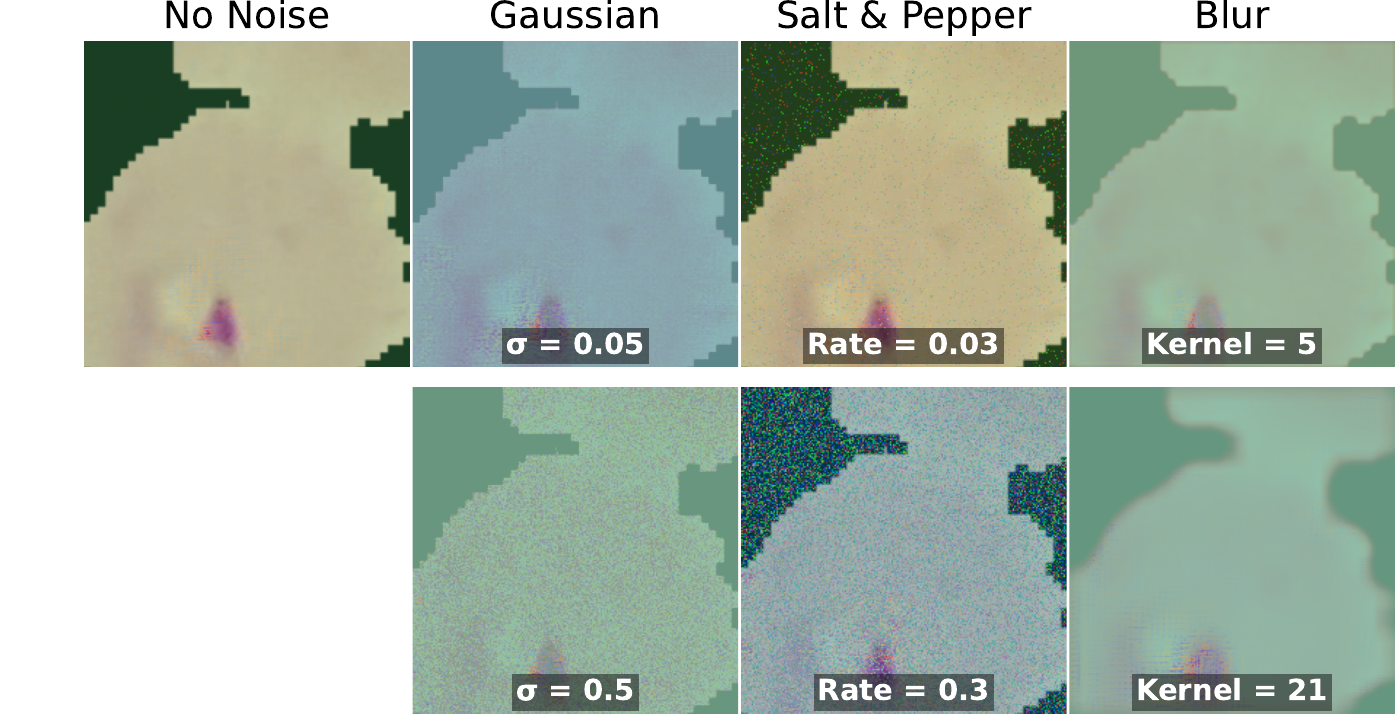}
            \caption{Integrated Gradients}
        \end{subfigure}
    
        \caption{Mobile Model}
    \end{figure}
    
    \begin{figure}[H]
        \centering
    
        \begin{subfigure}{\columnwidth}
            \centering
            \includegraphics[width=\linewidth]{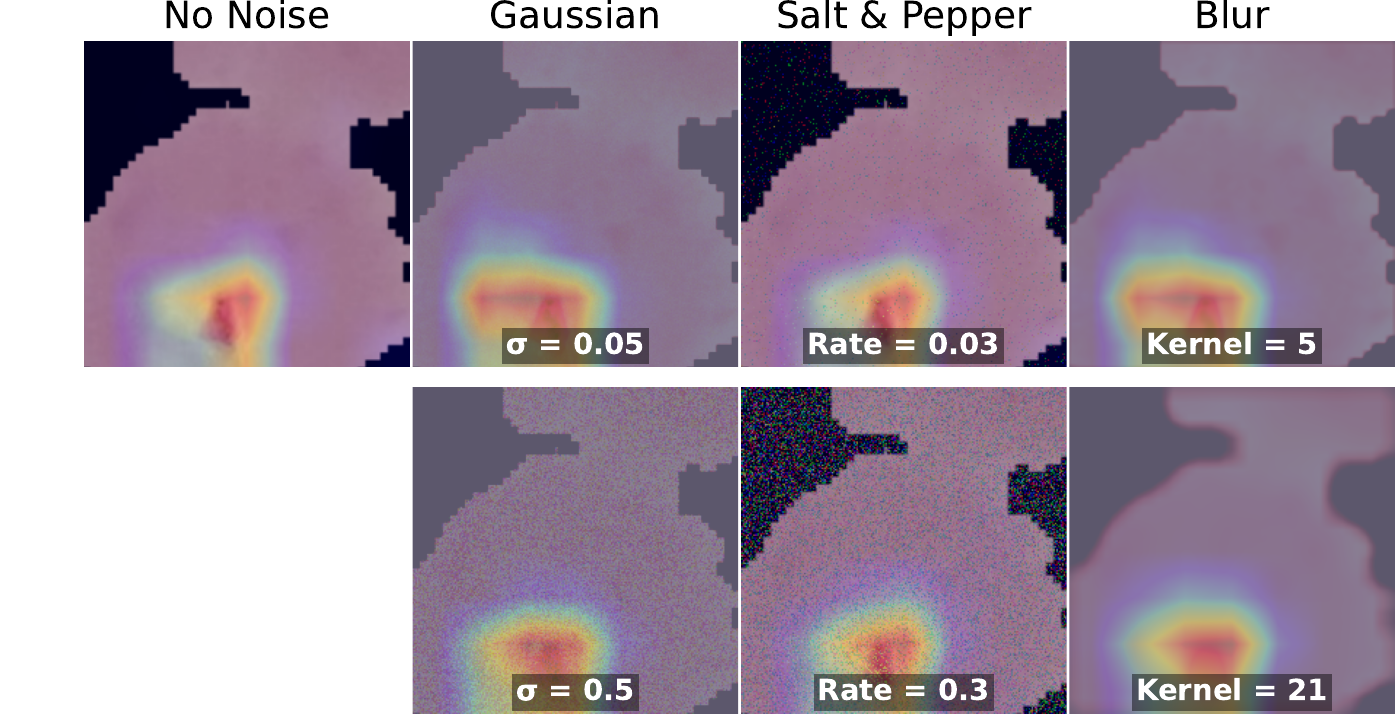}
            \caption{Grad-CAM}
        \end{subfigure}
        \hfill
        \begin{subfigure}{\columnwidth}
            \centering
            \includegraphics[width=\linewidth]{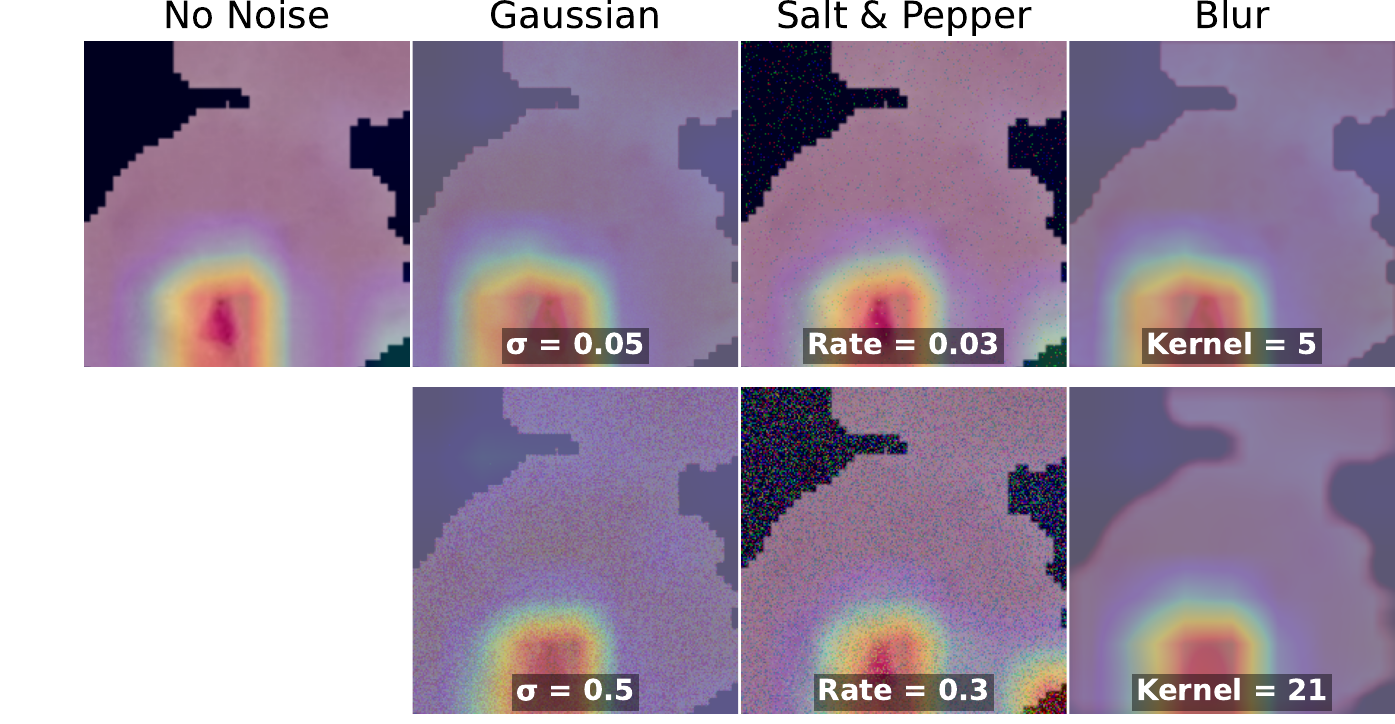}
            \caption{Score-CAM}
        \end{subfigure}
    
        \vspace{2mm}
    
        \begin{subfigure}{\columnwidth}
            \centering
            \includegraphics[width=\linewidth]{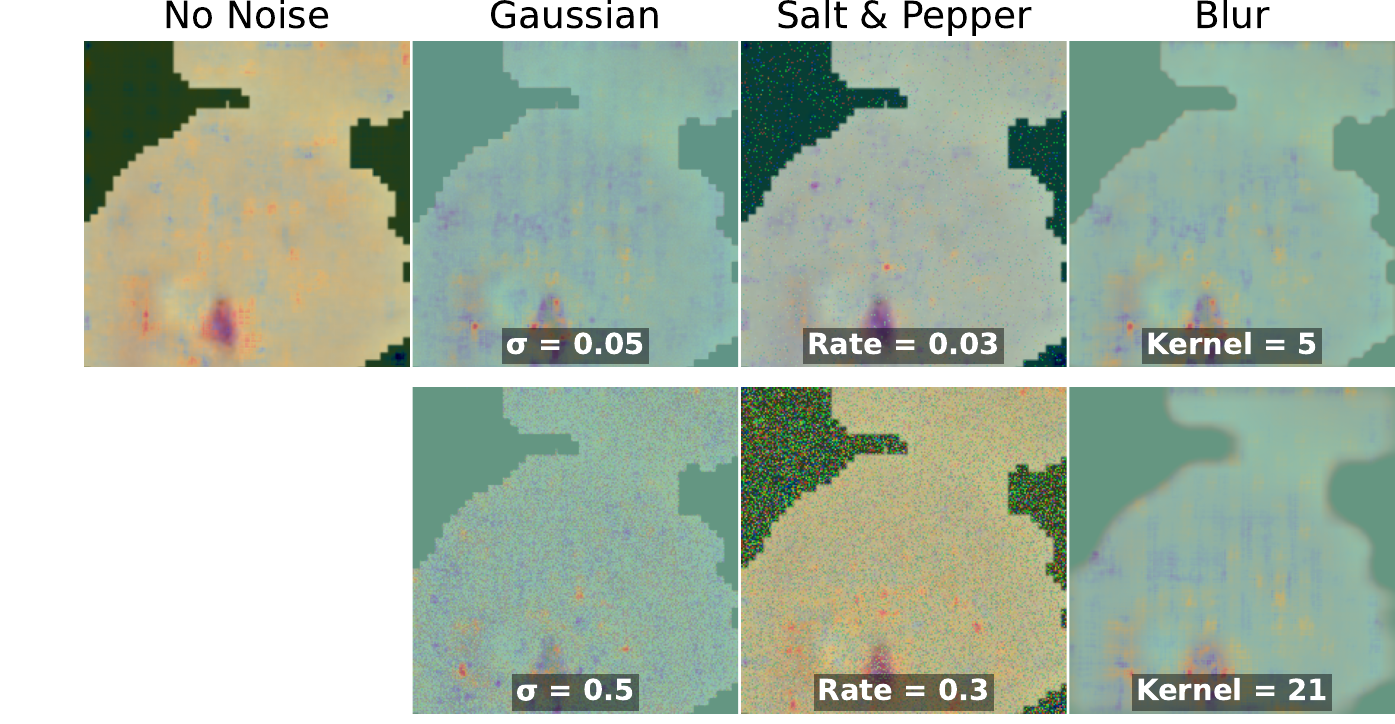}
            \caption{SHAP}
        \end{subfigure}
        \hfill
        \begin{subfigure}{\columnwidth}
            \centering
            \includegraphics[width=\linewidth]{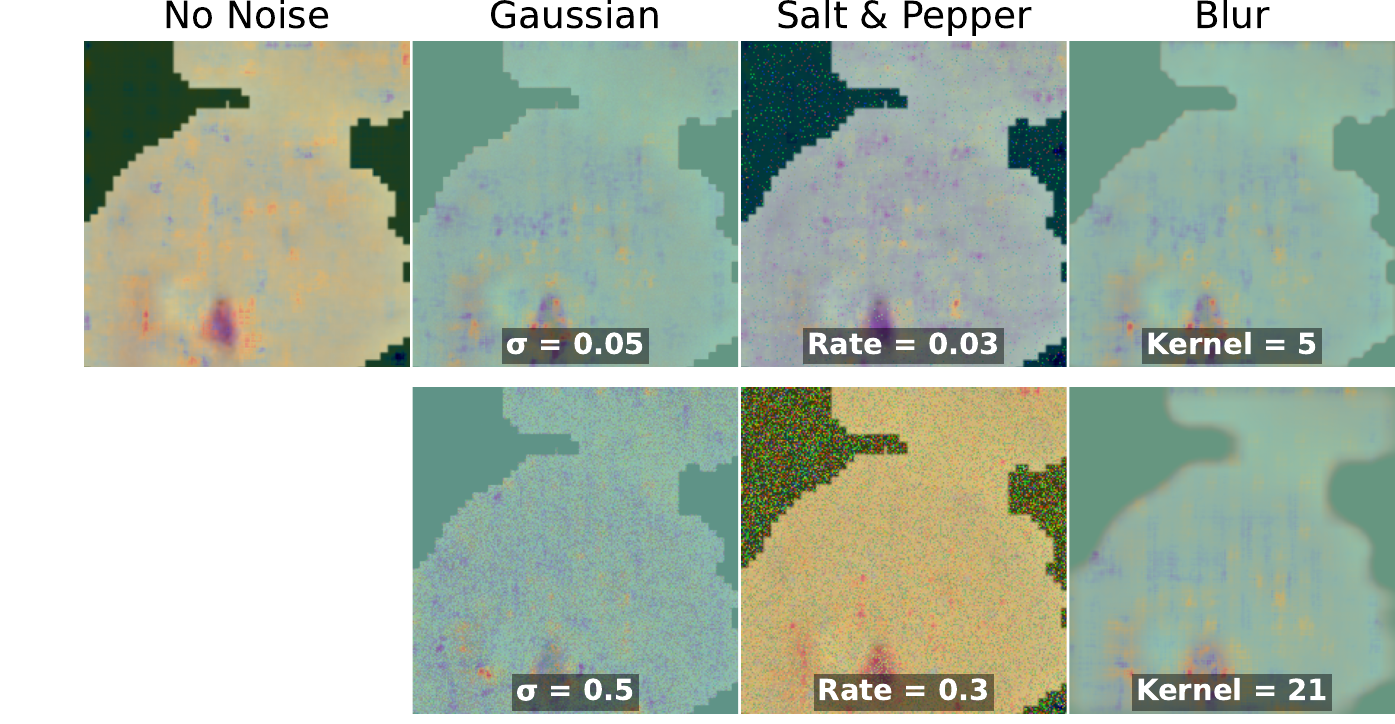}
            \caption{Integrated Gradients}
        \end{subfigure}
    
        \caption{ResNet Model}
    \end{figure}
    \begin{figure}[H]
        \centering
    
        \begin{subfigure}{\columnwidth}
            \centering
            \includegraphics[width=\linewidth]{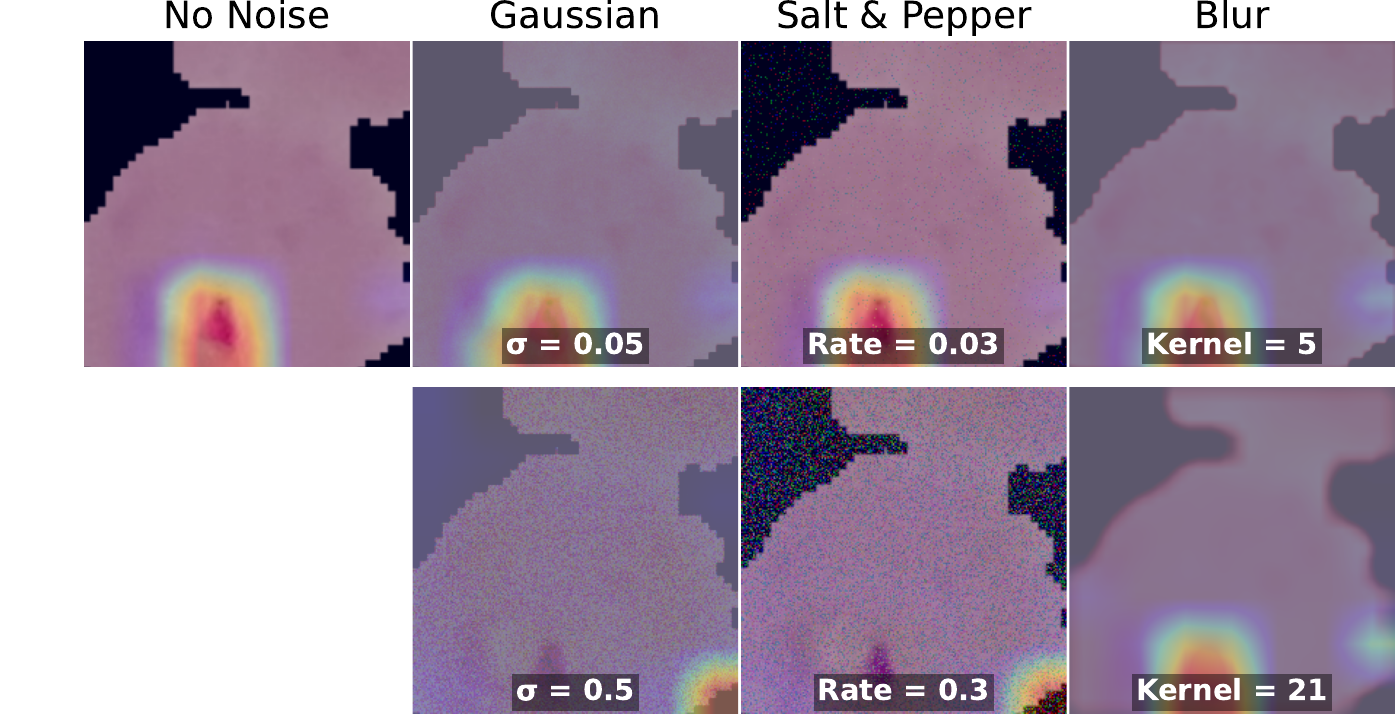}
            \caption{Grad-CAM}
        \end{subfigure}
        \hfill
        \begin{subfigure}{\columnwidth}
            \centering
            \includegraphics[width=\linewidth]{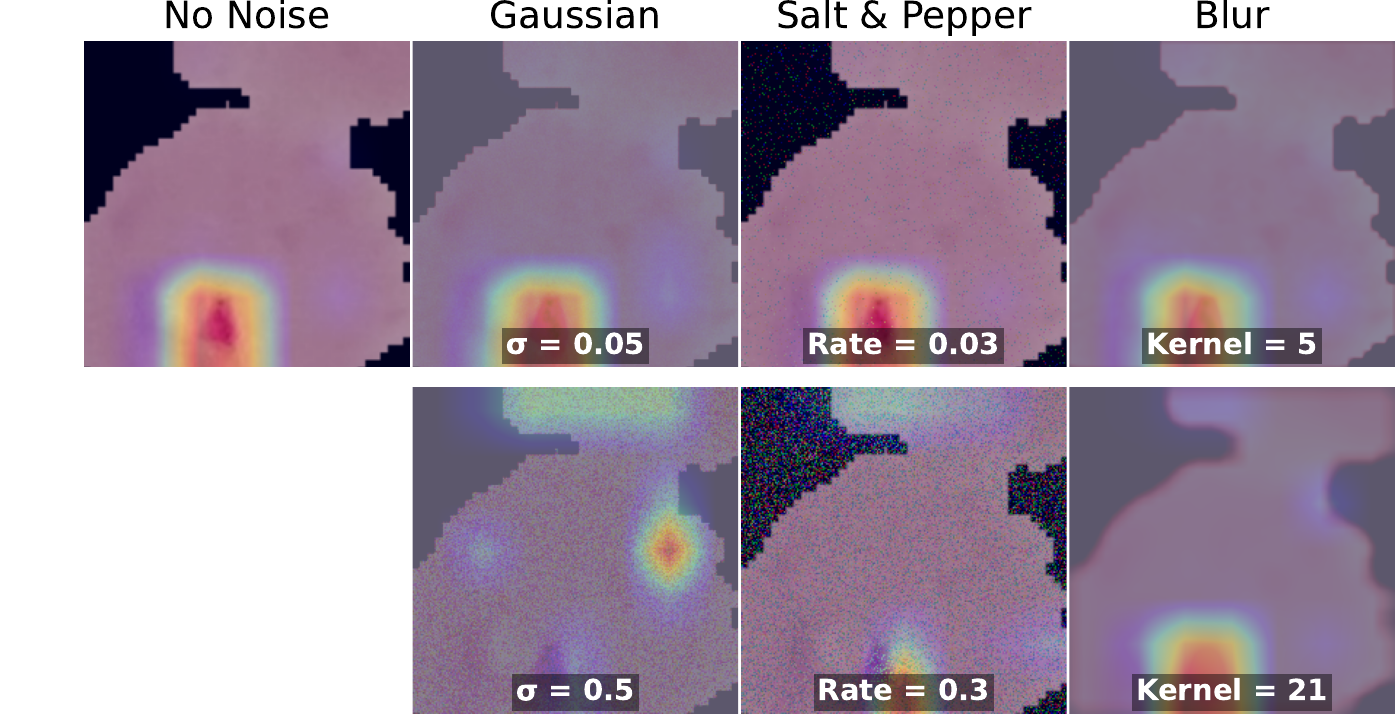}
            \caption{Score-CAM}
        \end{subfigure}
    
        \begin{subfigure}{\columnwidth}
            \centering
            \includegraphics[width=\linewidth]{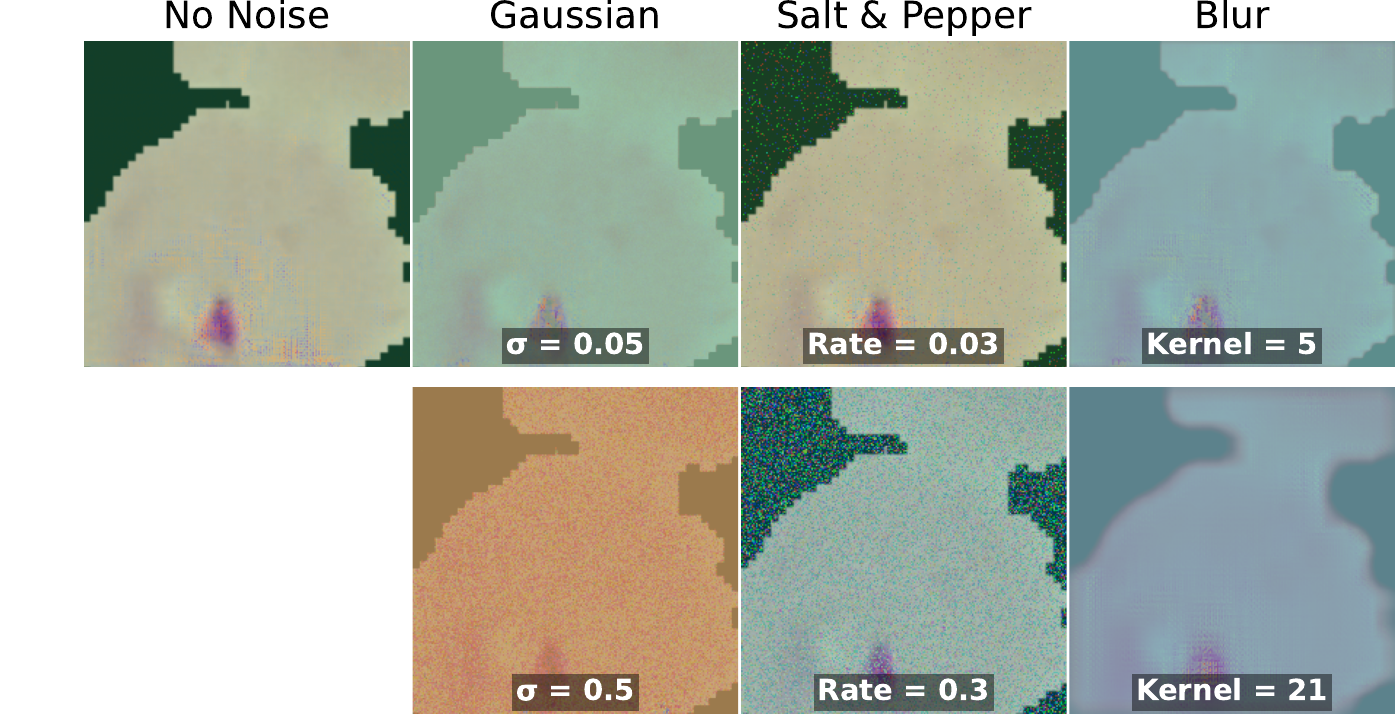}
            \caption{SHAP}
        \end{subfigure}
        \hfill
        \begin{subfigure}{\columnwidth}
            \centering
            \includegraphics[width=\linewidth]{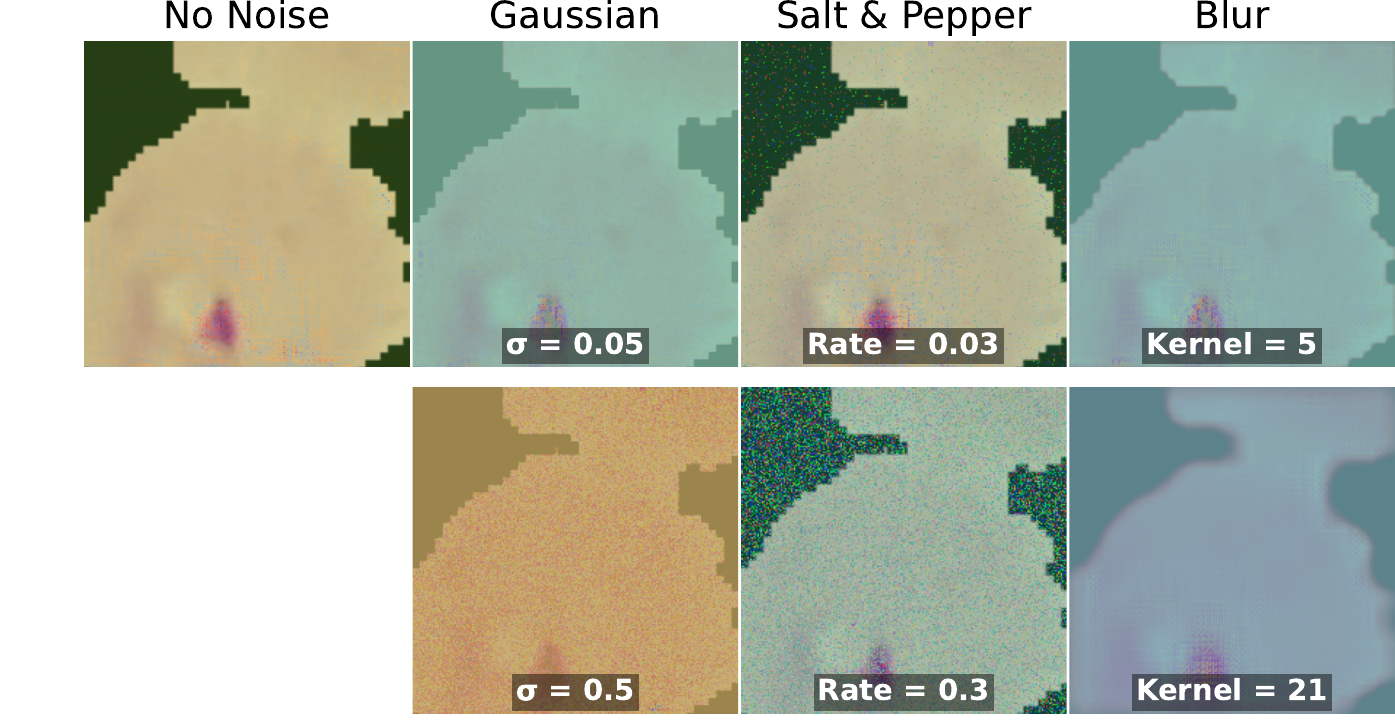}
            \caption{Integrated Gradients}
        \end{subfigure}
    
        \caption{EfficientNet Model}
    \end{figure}

\section{Insertion metric curves with random baseline}
    \label{baselinesinsertion}
    The insertion curve measures how quickly model confidence recovers as the most important patches -- as identified by each XAI method -- are progressively revealed from a fully masked mean baseline. At $x=0$, the model sees only the mean baseline image and confidence is near zero. At $x=100\%$, the full original image is restored. The blue (XAI) curve should therefore rise faster and sit above the red (random) curve throughout, which is the case.
    \begin{figure*}[t]
        \centering
        \includegraphics[width=\textwidth]{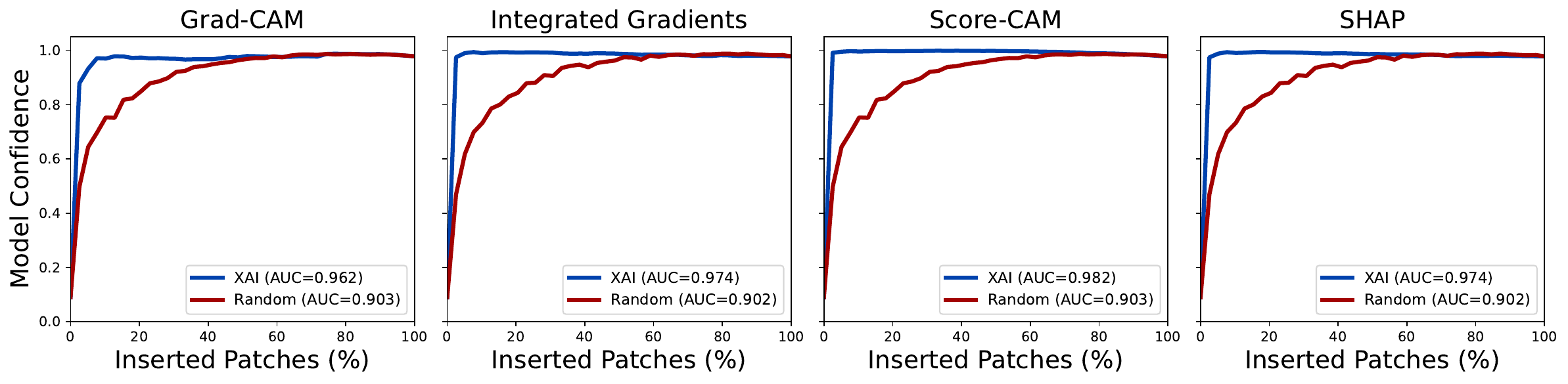}
        \caption{EfficientNet: insertion curves for different post-hoc XAI methods with random baseline.}
        \label{effnet_baseline}
    \end{figure*}
    
    \begin{figure*}[t]
        \centering
        \includegraphics[width=\textwidth]{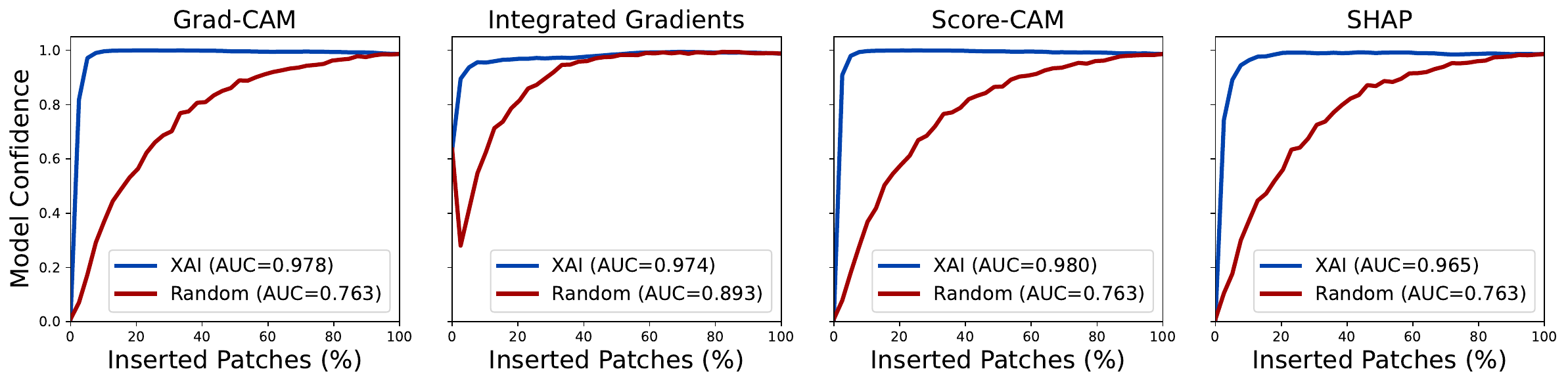}
        \caption{ResNet: insertion curves for different post-hoc XAI methods with random baseline.}
        \label{Resnet_baseline}
    \end{figure*}
    
    \begin{figure*}[t]
        \centering
        \includegraphics[width=\textwidth]{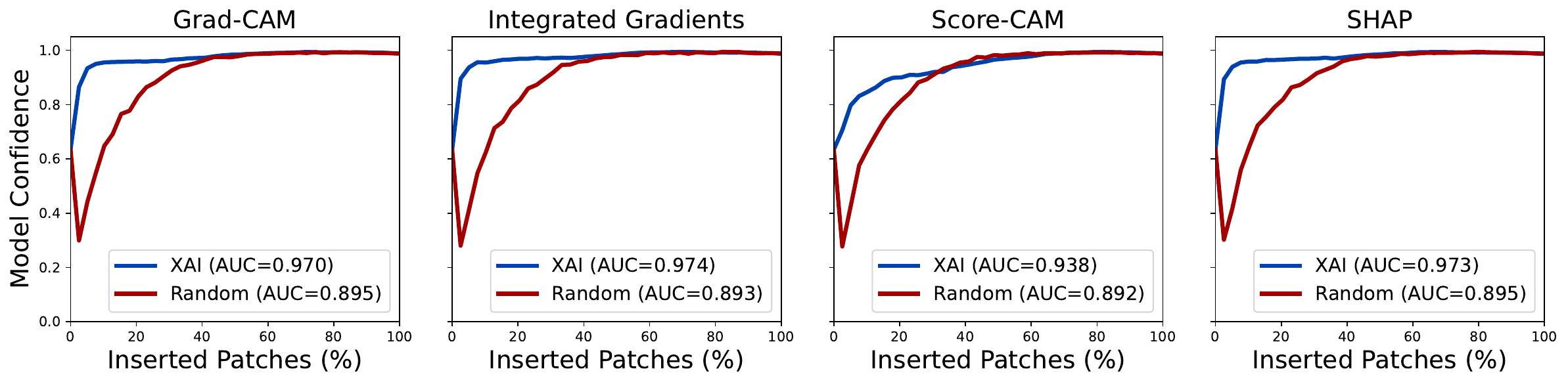}
        \caption{MobileNet: insertion curves for different post-hoc XAI methods with random baseline.}
        \label{mobilenet_baseline}
    \end{figure*}

\end{document}